\documentclass[10pt,twocolumn,letterpaper]{article}

\usepackage{iccv}
\usepackage{times}
\usepackage{epsfig}
\usepackage{graphicx}
\usepackage{amsmath}
\usepackage{amssymb}

\usepackage{booktabs}
\usepackage{tabularx}
\usepackage{multirow}
\usepackage{pifont}
\usepackage[export]{adjustbox}

\usepackage{float}
\usepackage{multicol}

\usepackage[pagebackref=true,breaklinks=true,letterpaper=true,colorlinks,bookmarks=false]{hyperref}

\iccvfinalcopy % *** Uncomment this line for the final submission

\def\iccvPaperID{1951} % *** Enter the ICCV Paper ID here

\ificcvfinal\fi

\begin{document}

%%%%%%%%% TITLE

\title{Examining Autoexposure for Challenging Scenes}
%0.3 hspace before so everything looks center on page
\author{SaiKiran Tedla\textsuperscript{*} \hspace{1.5cm}Beixuan Yang\textsuperscript{*}\hspace{1.5cm}Michael S. Brown \vspace{2.5mm} \\
York University\\
{\tt\small [tedlasai, beixuan, mbrown]@yorku.ca}
}

%\maketitle

\twocolumn[{
\maketitle
}]

\def\thefootnote{*}\footnotetext{Equal contribution.}
\ificcvfinal\thispagestyle{empty}\fi

%%%%%%%%% ABSTRACT
\begin{abstract}
Autoexposure (AE) is a critical step applied by camera systems to ensure properly exposed images. While current AE algorithms are effective in well-lit environments with constant illumination, these algorithms still struggle in environments with bright light sources or scenes with abrupt changes in lighting. A significant hurdle in developing new AE algorithms for challenging environments, especially those with time-varying lighting, is the lack of suitable image datasets. To address this issue, we have captured a new 4D exposure dataset that provides a large solution space (i.e., shutter speed range from \(\frac{1}{500}\) to 15 seconds) over a temporal sequence with moving objects, bright lights, and varying lighting. In addition, we have designed a software platform to allow AE algorithms to be used in a plug-and-play manner with the dataset. Our dataset and associate platform enable repeatable evaluation of different AE algorithms and provide a much-needed starting point to develop better AE methods. We examine several existing AE strategies using our dataset and show that most users prefer a simple saliency method for challenging lighting conditions.
\end{abstract}

%%%%%%%%% BODY TEXT
%%%%%%%%% BODY TEXT
\section{Introduction and Motivation}\label{sec:intro}

This paper is focused on auto exposure (AE) for scenes with challenging lighting. A camera's AE subsystem is responsible for determining capture-time settings to ensure a properly exposed image. Exposure is attributed to the lens aperture, shutter speed, and sensor ISO gain; however, it is often the case that some of these parameters are fixed at capture time. For example, this paper focuses on the most common form of AE, known as aperture-priority AE, where the aperture (and ISO) are fixed to avoid changes in the depth of field. In aperture-priority, AE algorithms are responsible for choosing the appropriate shutter speed based on information gleaned from the current captured image. AE is inherently a dynamic process, where the algorithm constantly predicts changes in shutter speed to account for a changing environment. While AE is a long-standing and well-researched problem in low-level computer vision, AE algorithms, including those on commercial cameras, still struggle in challenging lighting environments.

\begin{figure}
	\begin{center}
	\includegraphics[width=0.5\textwidth]{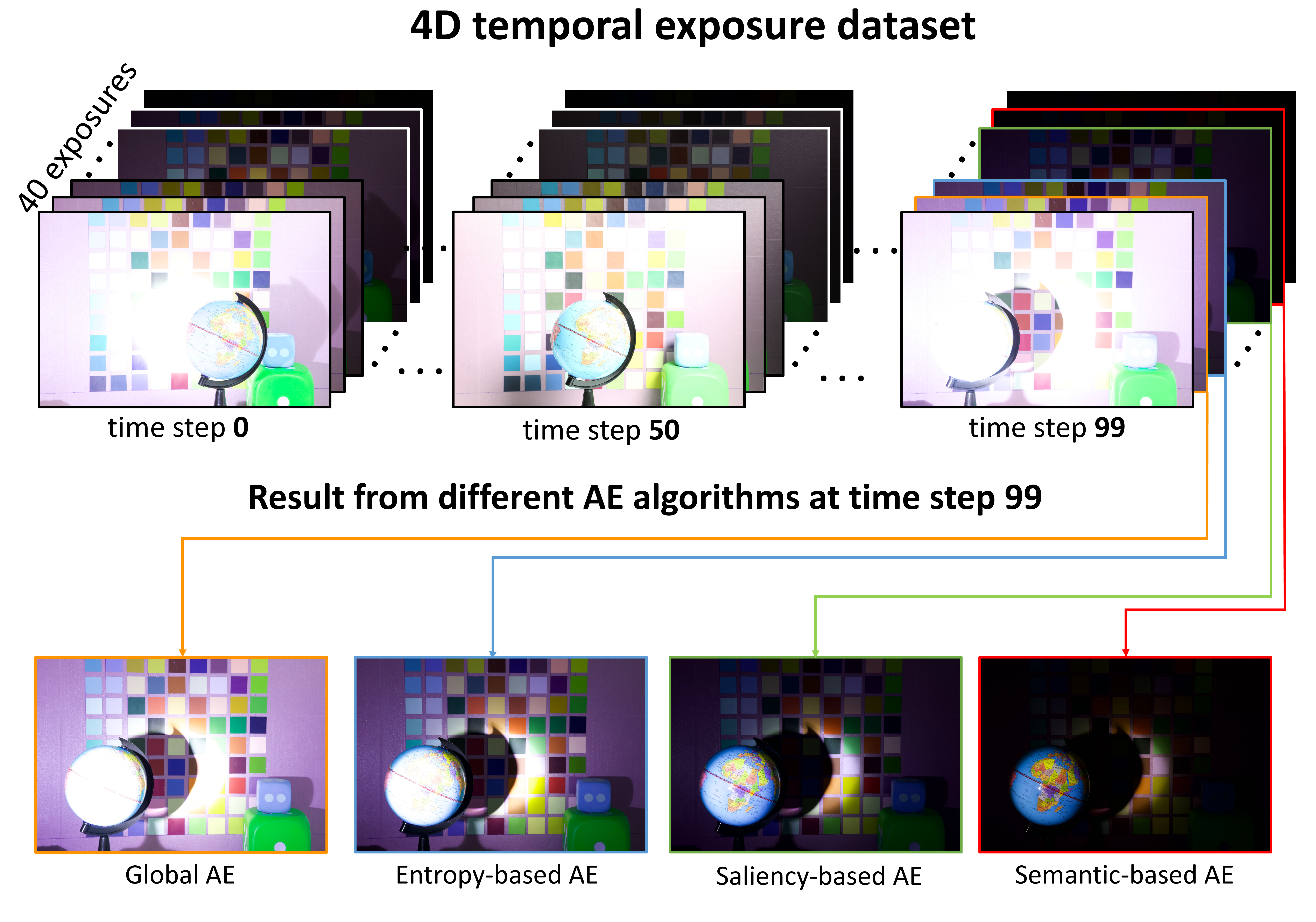}
	\end{center}
	\caption{An example scene from our new 4D exposure dataset. A moving object (globe) has a light source that is periodically turned on and off, causing an abrupt change in lighting. Using our dataset and AE platform, we can evaluate different AE algorithms. Here we show four different AE algorithms which select different exposures as their solution for a particular time step.}
	\label{fig_teaser}
\end{figure}

Two primary factors make scenes challenging for AE algorithms. The first is when the scene's dynamic range surpasses the capability of the imaging sensor. In such cases, the AE algorithm must determine which part of the scene will be over- or underexposed.   For this problem, it is easy to construct 3D datasets comprised of a stack of 2D images of a stationary scene captured under intense lighting with different shutter speeds. Several such exposure datasets exist, providing a complete solution space with fixed lighting and objects. The second challenging factor for AE is when there is an abrupt change in the environment lighting, such as lights switching on and off or objects moving under bright lights. Capturing datasets for this scenario is more complicated, given the temporal nature of the changing scene. To the best of our knowledge, there are no 4D datasets that provide a complete solution space over a dynamically changing environment.   This latter case is the impetus for our work. Specifically, we have captured a 4D dataset using a stop-motion setup.   Our dataset comprises nine (9) scenes, each with 100 time steps and each time step with 40 exposures. Figure~\ref{fig_teaser} shows an example of a scene in our dataset. Scenes are carefully constructed to cover a range of challenging conditions, including objects moving in front of intense lights, scenes with highly reflective materials, and scenes with rapid lighting changes. We have also developed a software platform to use with our dataset.

Our new dataset and platform allow us to evaluate various AE algorithms and compare their results on precisely the same input and starting conditions (e.g., starting in an over-exposed or under-exposed condition). Figure~\ref{fig_teaser} shows that different AE algorithms select notably different shutter speeds at different time steps. Given the mixed results of these methods, we performed a user study to determine the preferences of these different AE algorithms when used in challenging scenes. While our evaluation represents early work, most users preferred our AE algorithm based on a simple saliency method to determine regions of interest in the image. In the following, we detail the construction of our dataset, an API designed to allow AE algorithm testing, the details of the various AE algorithms tested, and the results from our user study.

\section{Related Work}

This section discusses AE algorithms and datasets.

\noindent{\textbf{AE methods.}}~
While manual exposure control is possible for still photography, robust AE is needed when capturing video or photographs in dynamic environments~\cite{personalized_auto_exposure,onzon2021neural}. Many AE algorithms are proprietary~\cite{proprietary_auto, propriety_auto_exposure_2} because camera manufacturers implement AE in hardware due to the need for low latency~\cite{onzon2021neural}. AE algorithms in the literature fall into two broad categories: (1) content-agnostic AE and (2) semantic AE.

Content-agnostic AE approaches do not explicitly consider the scene content but instead examine heuristics derived from images to determine the appropriate shutter speed ~\cite{battiato2010image, schulz2007using, nasser2002, surveillance2015}. These methods typically examine the global mean of the image's histogram to determine a change in the shutter speed, resulting in the current histogram mean being changed to some target value~\cite{kuno1998, nasser2002, histogramMean}. Many AE methods are variants on this basic idea but modify how the histogram is constructed to avoid pixels in over-exposed and under-exposed regions contributing to the histogram~\cite{kuno1998, multiexposurefusion, cho1999, VuongANA}. Many works also treat AE as a model-predictive control problem with the goal of fast correction after a poor exposure, based on histogram heuristics~\cite{mpc1, mpc2, mpc3}.

Another common approach for content-agnostic AE is to measure optimal exposure based on entropy instead of simple histogram moments. Zhang et al.~\cite{entropy3} proposed a method where the best exposure was the shutter speed that maximized the entropy of the histogram. This approach requires capturing multiple exposures to find the maximum entropy. Subsequent works attempt to reduce the search time~\cite{entropy1, entropy2} and settle for local maximums in entropy.

Semantic AE methods explicitly consider scene content by weighting different image regions related to their semantics. Many consumer cameras emphasize scenes with faces~\cite{face_auto_exposure}. Yang et al.~\cite{personalized_auto_exposure} apply reinforcement learning to create personalized semantic AE based on user preference. Onzon et al.~\cite{onzon2021neural} recently proposed a method to train semantic AE jointly with the task of object recognition. This method is ideal for machine vision but is not optimized for perceptual quality. To this point, it is worth noting that many methods in the computer vision literature process poorly exposed images to improve the perceptual quality of a capture image~\cite{Yuan2012AutomaticEC,postprocess1, postprocess2, postprocess3, postprocess4}. However, these post-processing methods do not directly improve the basic AE algorithm used to capture the image in the first place.

\noindent{\textbf{AE Datasets.}}~Several existing datasets have been collected for tasks related to exposure-related problems. Most of these datasets contain only a single static scene with varying exposure, often with the goal of post-processing exposure correction~\cite{Dale2009,afifi2021,Guo_2020_CVPR}. Similarly, datasets have been captured with varying exposure to construct a fused HDR image~\cite{kwon2020,liu2020single,EKDMU17,Funt2010}. Recent work by ~\cite{Morawski2022} provided a dataset with intensive lighting for evaluating object detection and classification. While these can be used to evaluate AE on a static scene, none of these datasets include a temporal dimension.

Video datasets with multiple exposures to examine video-HDR methods have been captured~\cite{FroehlichSPIE2014,EKDMU17,chen2021hdr,6928652}. These existing datasets, however, often have limited exposure sampling (often only two exposures) and do not include scenes with harsh lighting or abrupt changes in lighting.  Moreover, AE algorithms are implemented in the low-level camera hardware and are therefore applied directly to RAW sensor data.  As a result, existing static and video-based datasets are unsuitable for evaluating AE algorithms.

As discussed in Section~\ref{sec:intro}, the lack of sufficient AE datasets is the impetus of our work.  The following sections describe our dataset collection and AE evaluation platform.  Representative content-agnostic and semantic AE algorithms are evaluated using our system.  Our dataset and platform even allow us to propose a simple semantic algorithm using fast saliency detection that shows promising performance.

\begin{figure*}[ht]
  \begin{center}
  \includegraphics[width=1\textwidth]{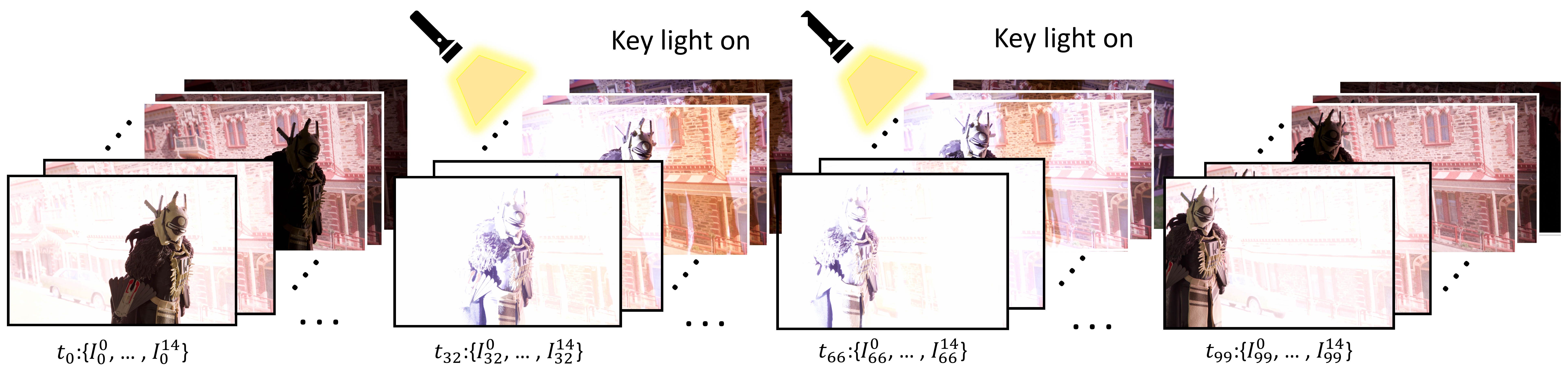}
  \end{center}
  \caption{4D Exposure Dataset (Scene 9). The four dimensions refer to (1) time, (2) exposure,
  and (3\&4) the 2D coordinates of the RAW sensor image associated with each
  (1) and (2). Each scene is characterized by a combination of challenging lighting conditions and objects. In this scene, a flashing light source is simulated by turning on the spot light between $t_{20}$-$t_{39}$ and $t_{60}$-$t_{79}$.}
  \label{fig_dataset}
  \end{figure*}

\begin{figure}[b]
	\begin{center}
	\includegraphics[width=0.45\textwidth]{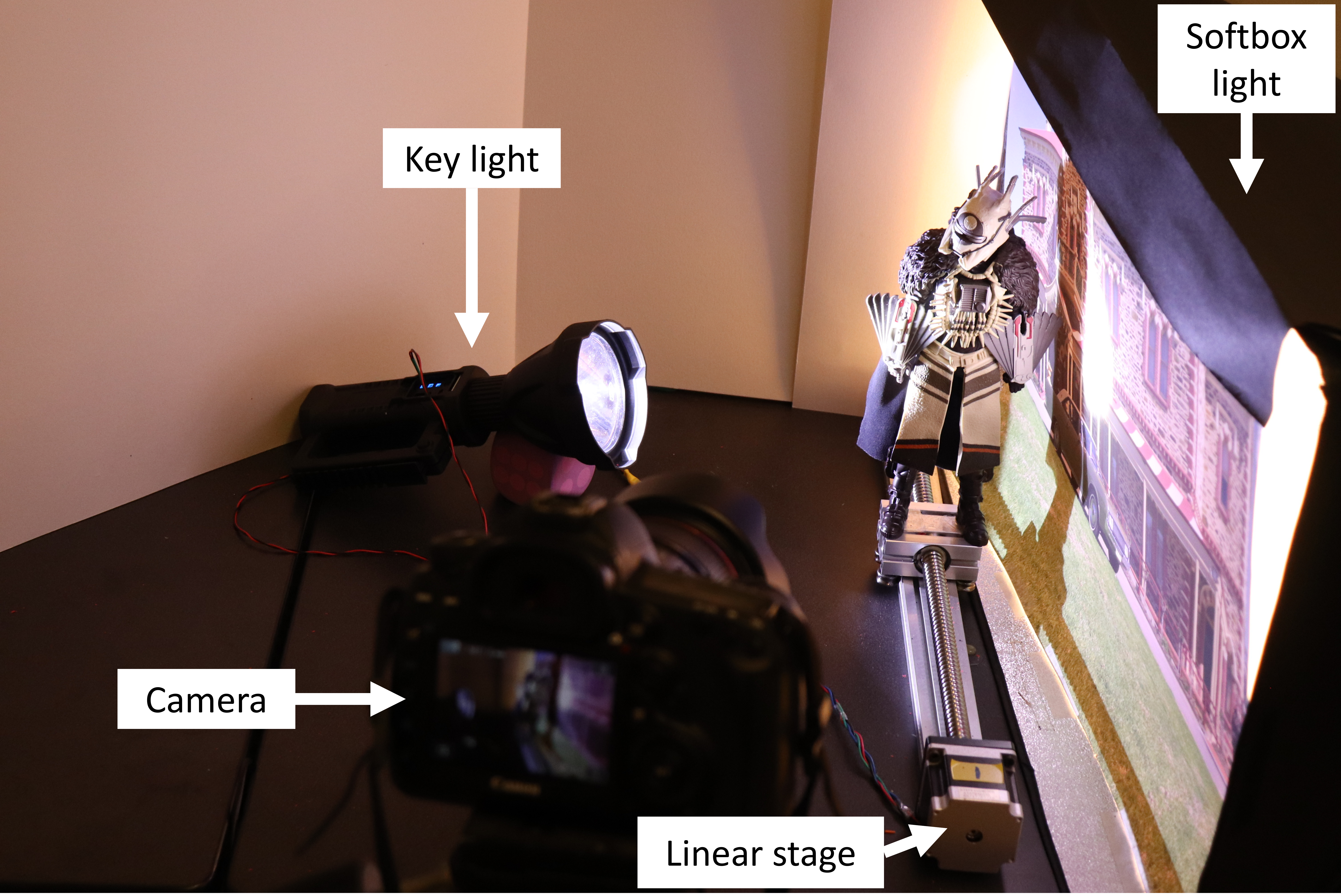}
	\end{center}
	\caption{Our dataset capture setup with controllable lights, camera, and motion stage.}
	\label{fig_setup}
\end{figure}
\section{4D Temporal Exposure Stack Dataset}
\label{sec:dataset}
%Michael-We say 4 dimensions but then mention 5, seems wierd....(Might be ok because the 4 dimensions are italicized...)
% Let's drop the 3.
We captured a temporal exposure dataset with four dimensions: $time \times exposure \times height \times width$. Our capture environment was adopted from the stop-motion setup proposed by Abuolaim et al.~\cite{Abuolaim2018} for 4D auto-focus capture (see Figure~\ref{fig_setup} for an image of our setup). All scenes were captured in a dark room with two controllable 90000-lumen spot lights and a softbox-diffused light source. The lights use DC power to prevent the flickering effect caused by alternating currents. The softbox light was used as a constant illumination while the spot lights provide intense lighting (high dynamic range) and can be turned on and off. Two linear stage actuators allow us to move objects and spot lights. An Arduino/Genuino microcontroller synchronized scene motion, lights, and camera capture.  Images were captured with a Canon EOS 5D Mark IV in RAW format (6720$\times$4480). Since we are interested in aperture-priority AE, we fixed the camera aperture (f/14) and ISO (100). %with varying exposure times. 

Our setup enabled us to capture scenes with dynamic lighting but have access to a full exposure stack at each time step. Each scene contained 100 simulated time steps $t_0-t_{99}$. For each time step $t$, we captured 15 exposure stack images $I^0_t-I^{14}_t$. The shutter speed ranged from \(\frac{1}{500}\) and 15 seconds---that is 12.87 EV of steps. We added interesting scene dynamics by turning on spot lights for certain time steps of our scenes. Scene 9 is shown in Figure~\ref{fig_dataset}.  In this scene, the spot light is on between $t_{20}-t_{39}$ and $t_{60}-t_{79}$, simulating sudden change in lighting. 

Our initial AE evaluation found that 15  exposure steps did not provide sufficient granularity for smooth exposure changes. Instead, we found that 40 exposure levels evenly sampled between \(\frac{1}{500}\) to 15 seconds provided a better emulation of exposure adjustment in a real AE system. To expand the initial 15 exposures to 40, any exposure not already part of the original 15 exposures is interpolated based on its two nearest neighboring exposure images in the original sequence. The interpolation procedure assumes a linear relationship between the exposure time and the image pixel value~\cite{mpc3}. To visualize the RAW images, we also process the RAW images to have a corresponding 4D sRGB dataset (also $100\times 40 \times height \times width\times3$) using the camera pipeline provided in~\cite{Abdelhamed2018}. 

In total, we captured nine scenes that provide a range of challenging setups for AE algorithms. Each scene had some combination of challenging lighting conditions: backlight, moving light, and/or sudden changes in lighting. In addition, some scenes contained reflective objects (i.e., a mirror) or preferred objects (i.e., faces). Table~\ref{table:scene_settings} shows each scene and the corresponding lighting conditions and objects.   

 \begin{table*}[]
    \centering
    \newcommand{\cl}{\centering\arraybackslash}
    \newcommand{\vc}[1]{\raisebox{-.5\height}{#1}}
    \newcommand{\cmark}{\ding{51}}%
    \newcommand{\xmark}{\ding{55}}%

    \setlength\tabcolsep{1pt}

    \begin{tabular}{ | >{\cl}p{17mm} | >{\cl}p{16mm} | >{\cl}p{16mm} | >{\cl}p{16mm} | >{\cl}p{16mm} | >{\cl}p{16mm} | >{\cl}p{16mm} | >{\cl}p{16mm} | >{\cl}p{16mm} | >{\cl}p{16mm} |}
      \hline
      Scene ID &  1 &   2 &  3 &  4 &  5 &   6 &  7 &  8 &  9 \\ \hline
      Example image & \multirow{2}{*}{\vc{\includegraphics[width=16mm,height=9.8mm,valign=M]{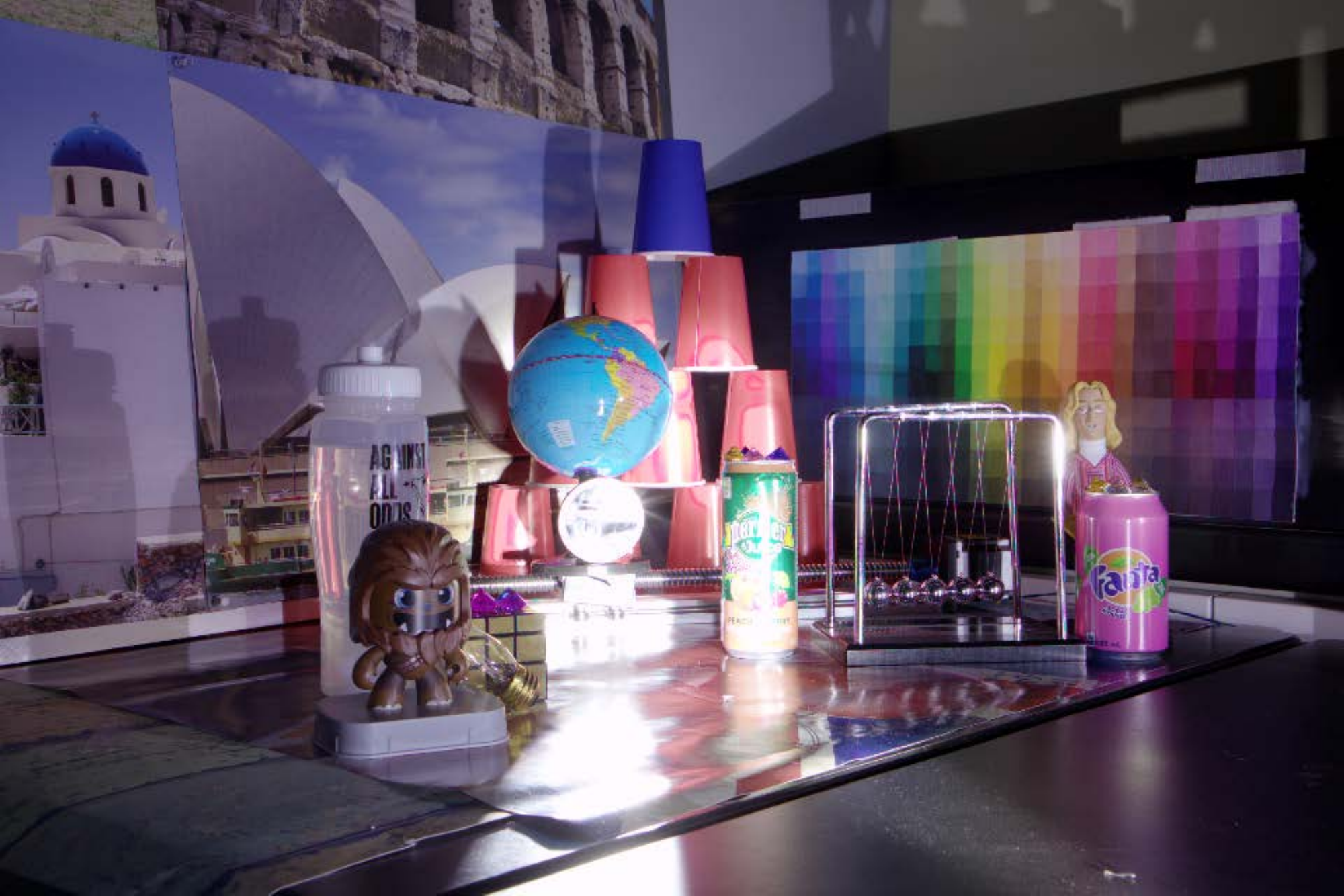}}} &
      \multirow{2}{*}{\vc{\includegraphics[width=16mm,height=9.8mm,valign=M]{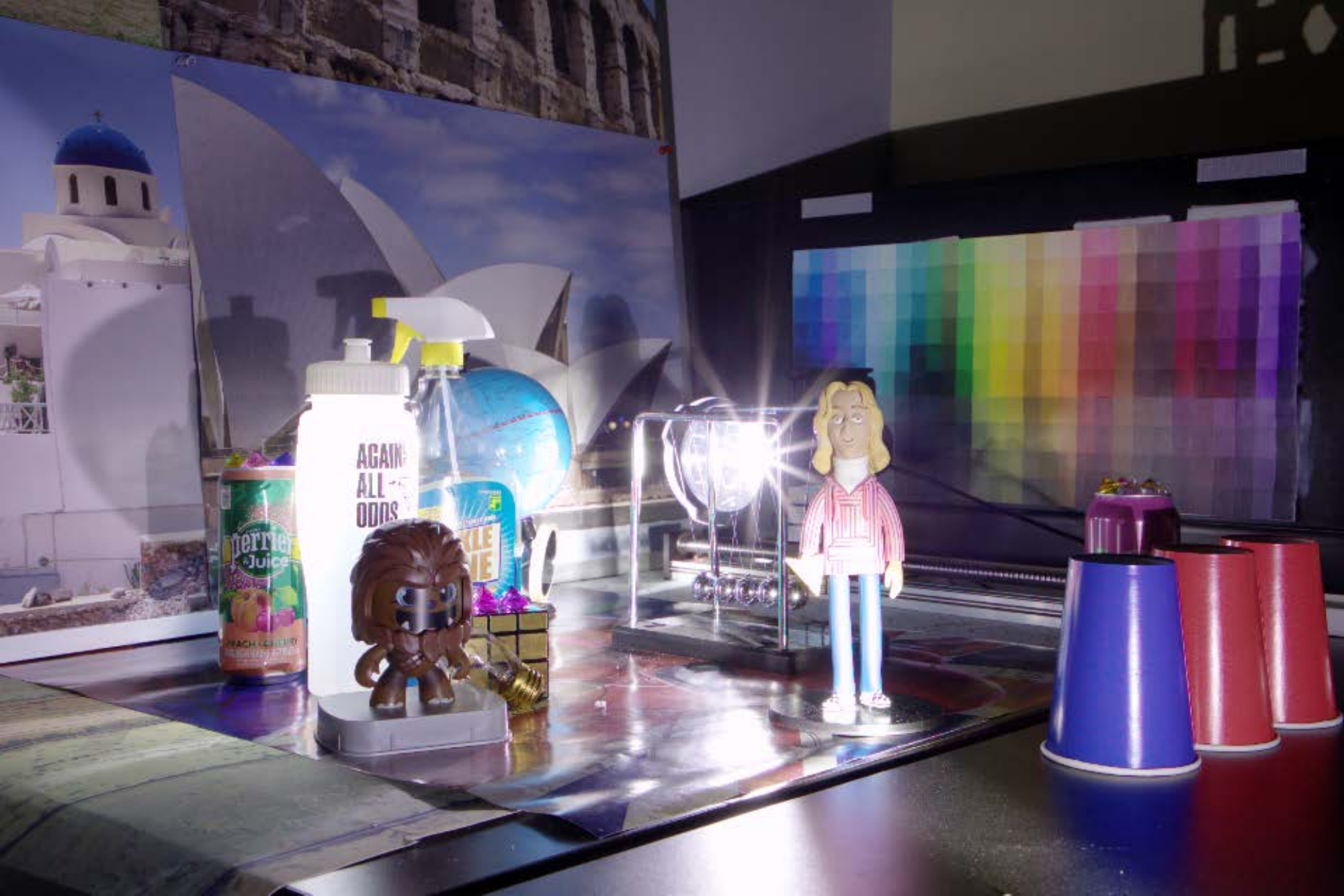}}} &
      \multirow{2}{*}{\vc{\includegraphics[width=16mm,height=9.8mm,valign=M]{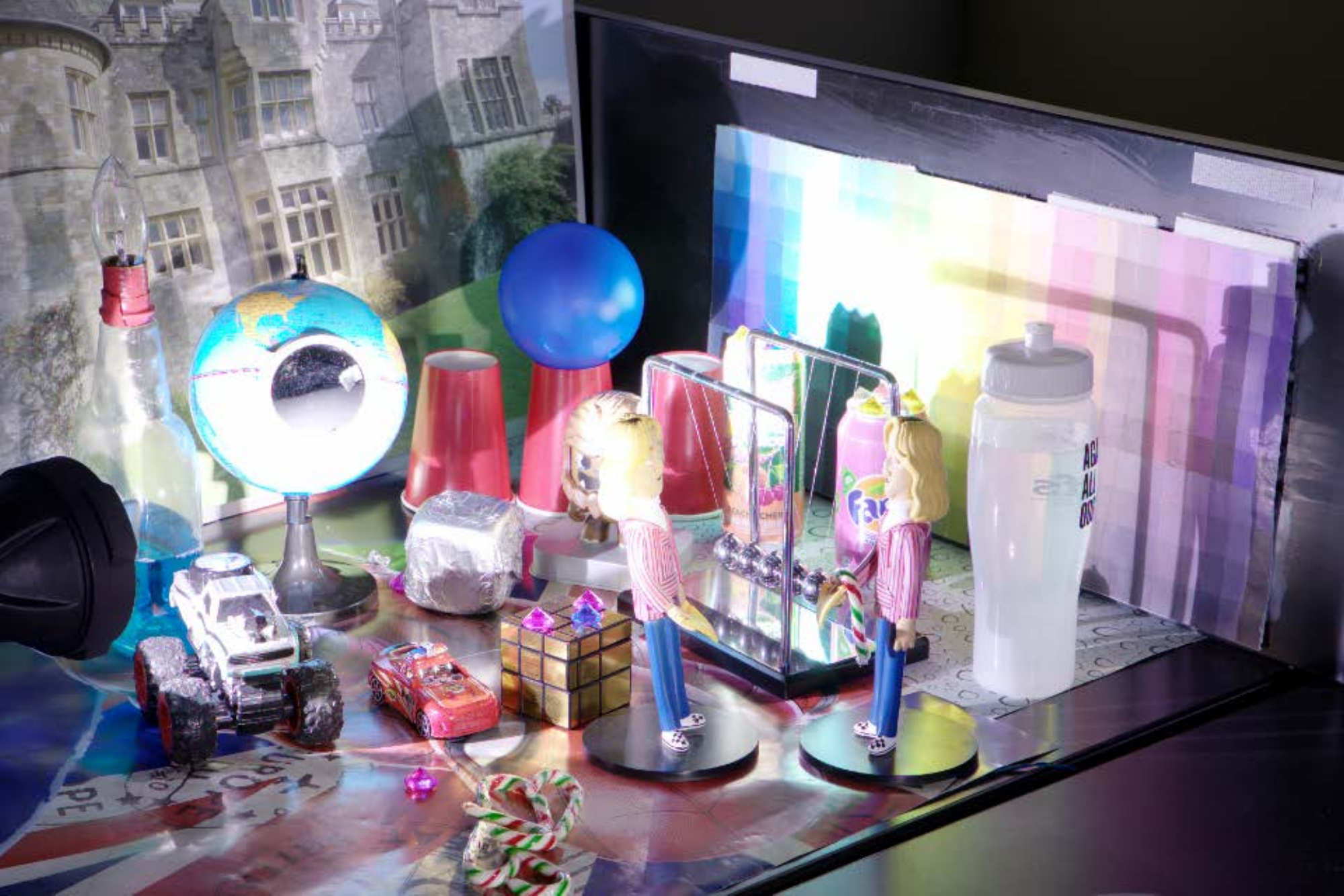}}} &
      \multirow{2}{*}{\vc{\includegraphics[width=16mm,height=9.8mm,valign=M]{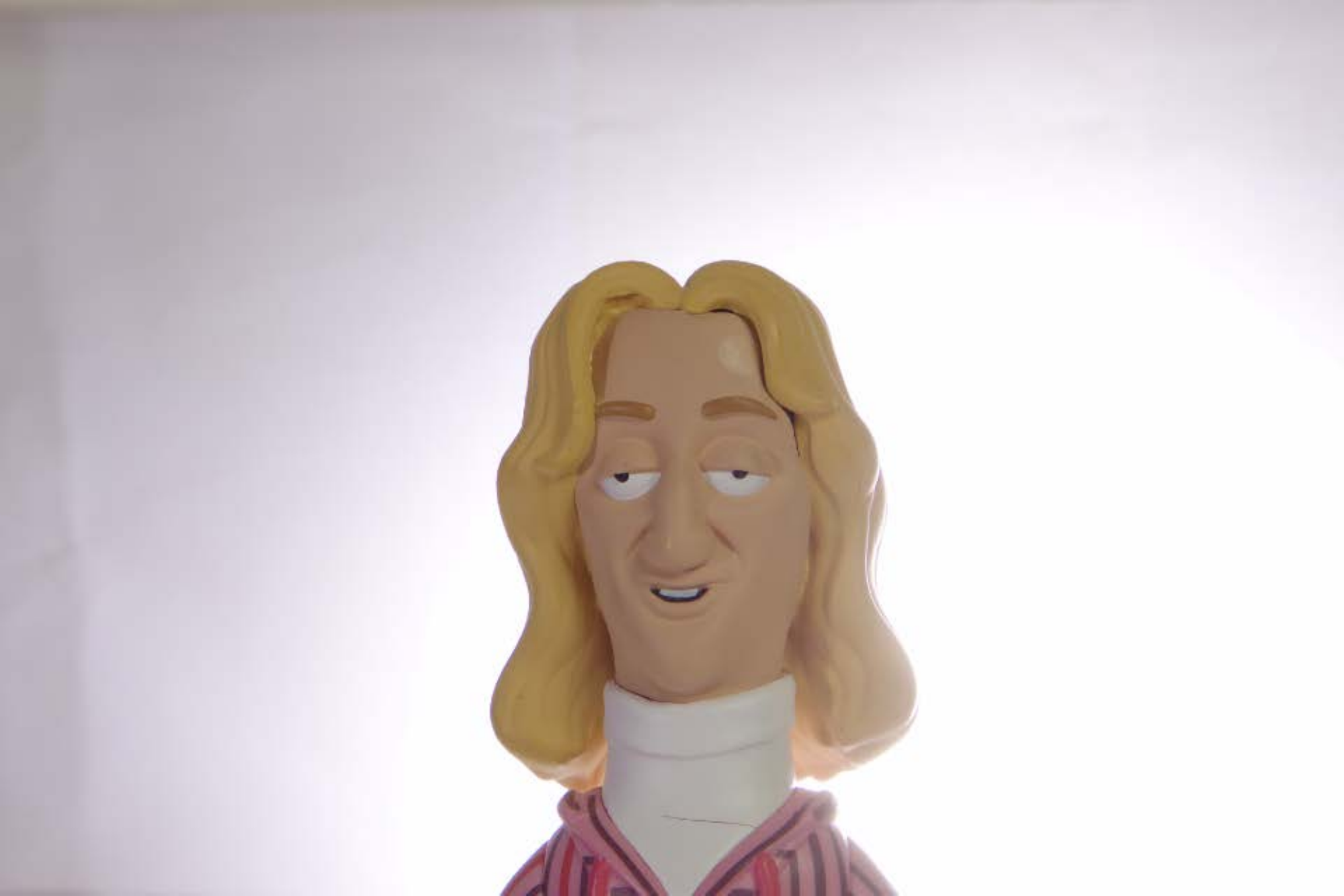}}} &
      \multirow{2}{*}{\vc{\includegraphics[width=16mm,height=9.8mm,valign=M]{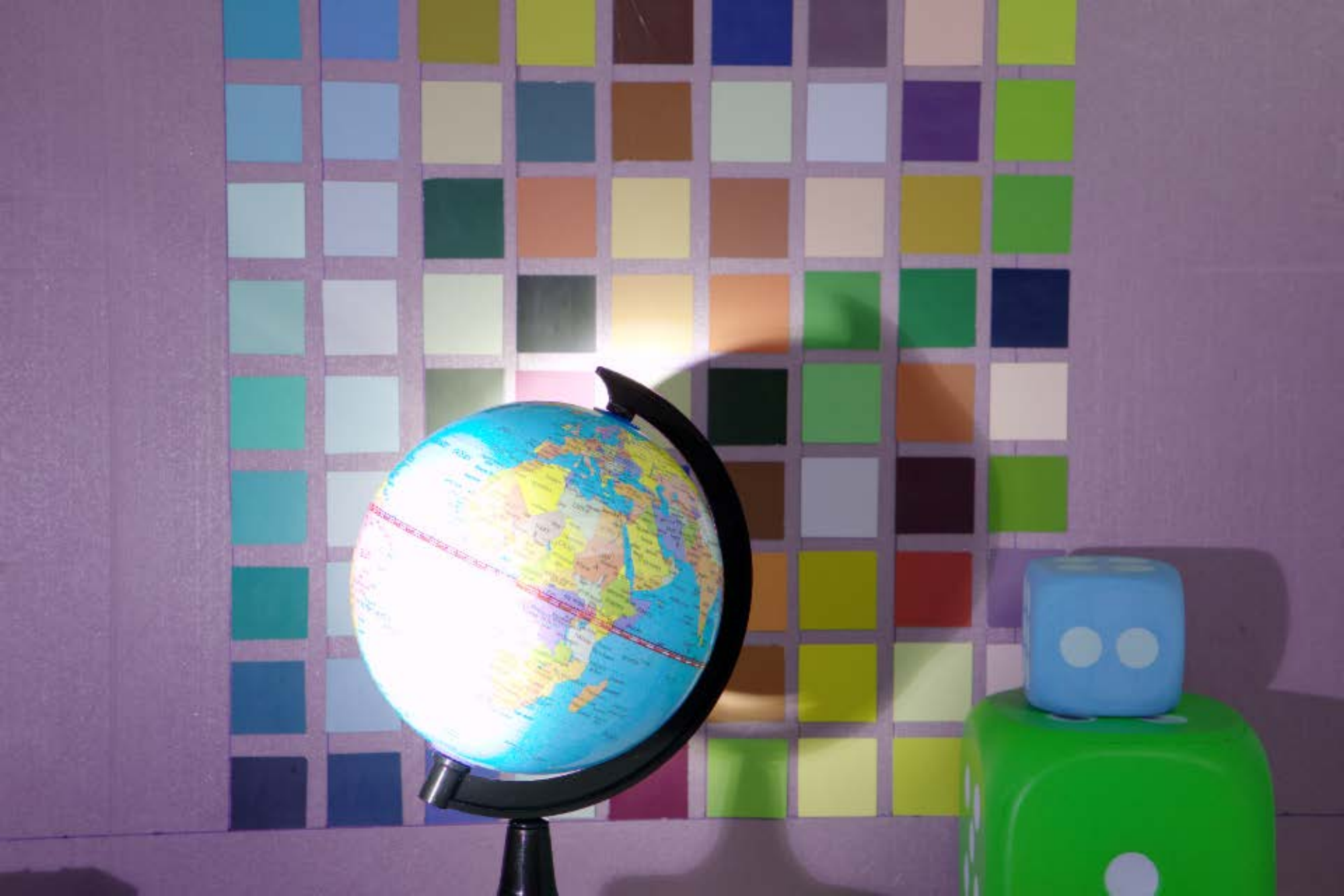}}} &
      \multirow{2}{*}{\vc{\includegraphics[width=16mm,height=9.8mm,valign=M]{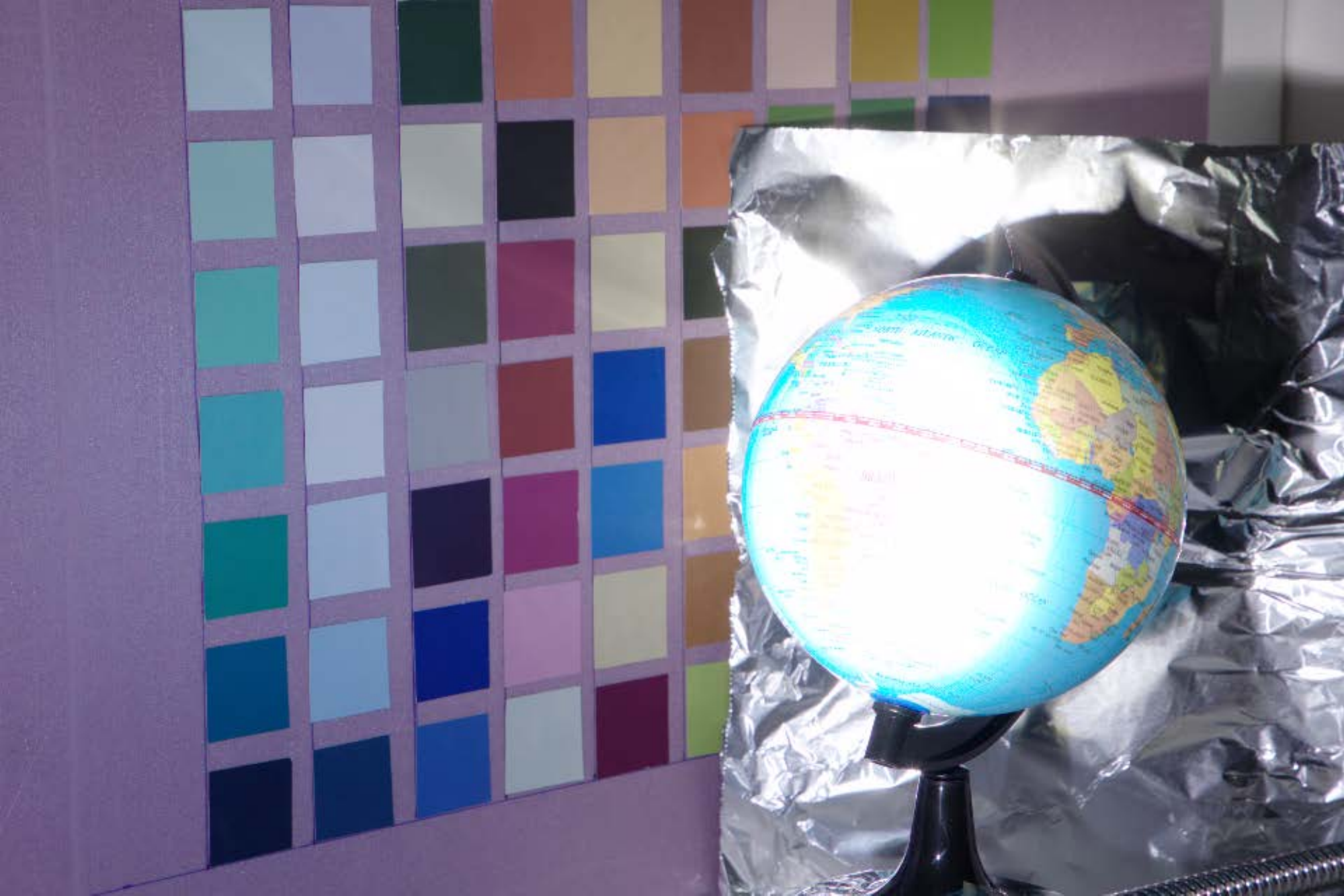}}} &
      \multirow{2}{*}{\vc{\includegraphics[width=16mm,height=9.8mm,valign=M]{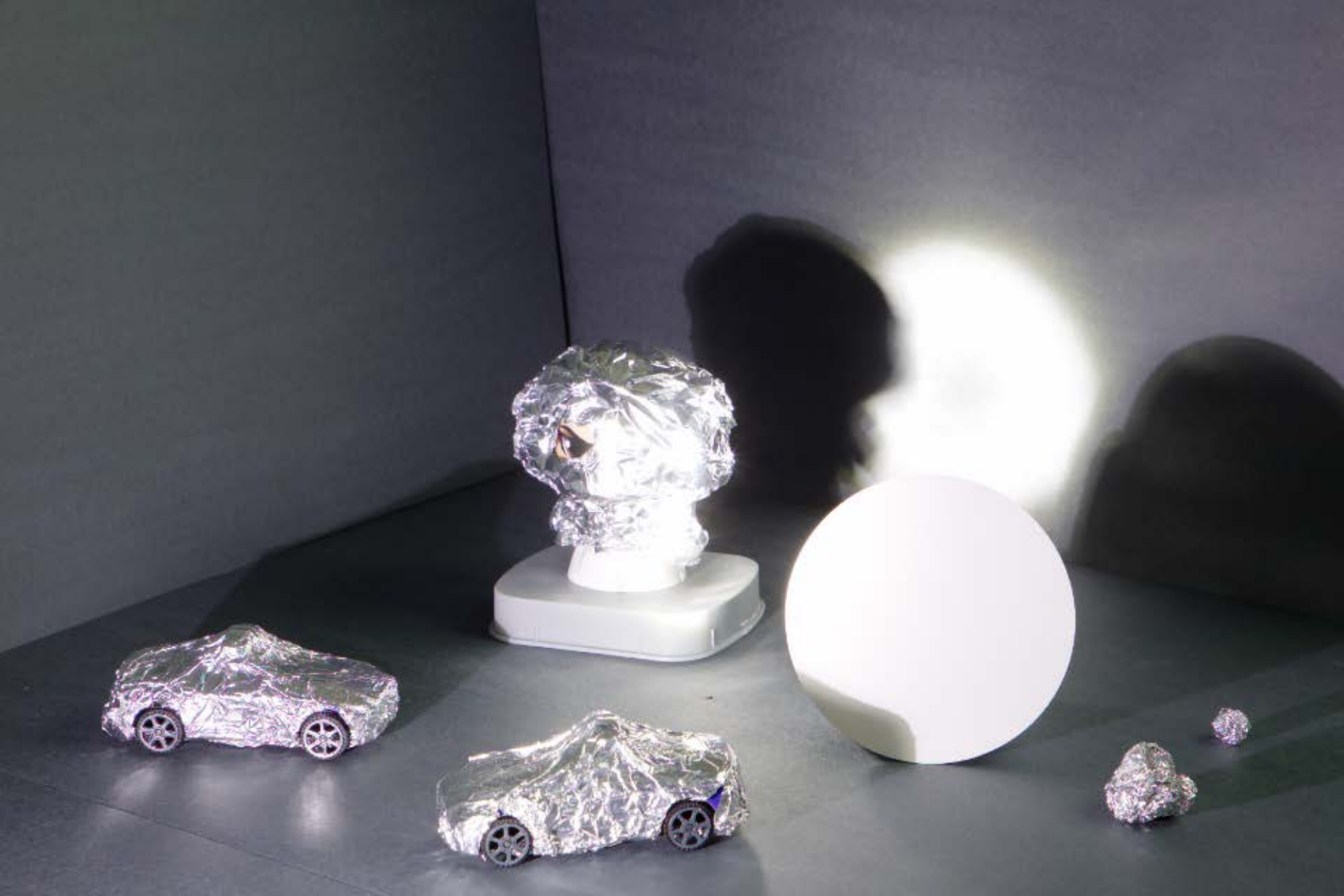}}} &
      \multirow{2}{*}{\vc{\includegraphics[width=16mm,height=9.8mm,valign=M]{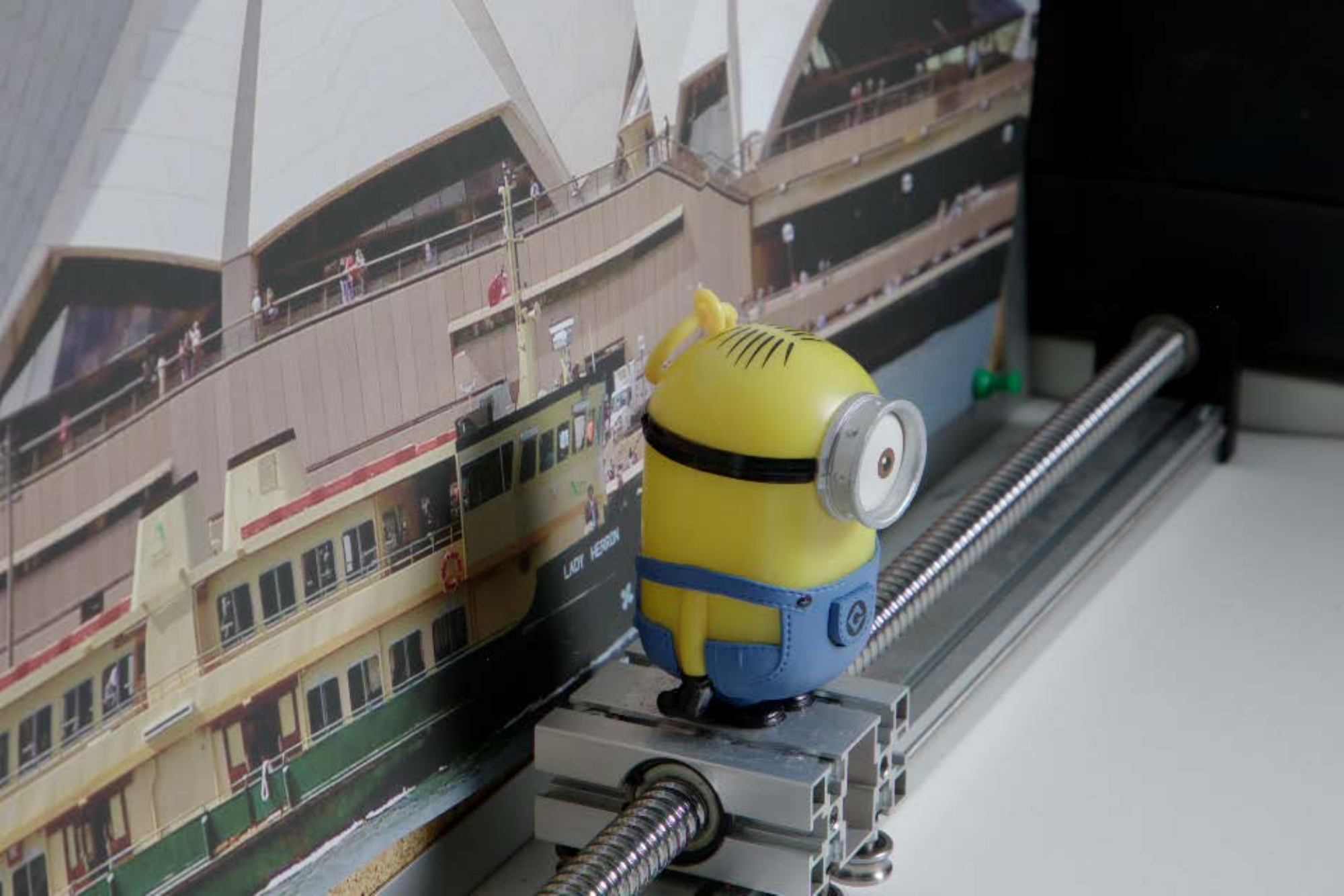}}} &
      \multirow{2}{*}{\vc{\includegraphics[width=16mm,height=9.8mm,valign=M]{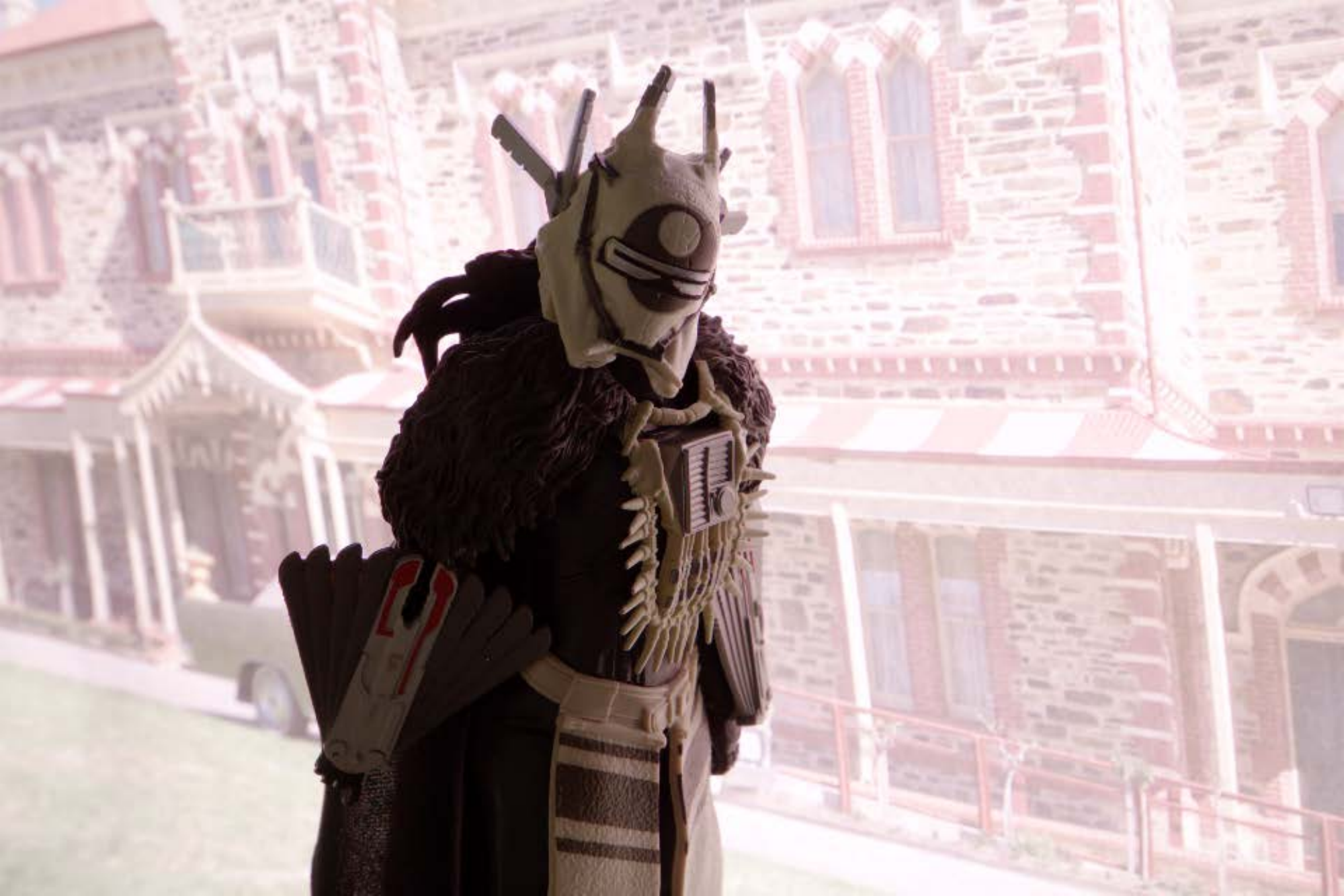}}} \\[6.5mm] \hline
      Back & \multicolumn{3}{c |}{\multirow{2}{*}{\xmark}} & \multirow{2}{*}{\cmark} & \multicolumn{4}{c |}{\multirow{2}{*}{\xmark}}&\multirow{2}{*}{\cmark}\\
      light & \multicolumn{3}{c |}{} &  & \multicolumn{4}{c |}{}&\\ \hline
      Moving light & \multirow{2}{*}{\xmark}&  \multicolumn{3}{c |}{\multirow{2}{*}{\cmark}}&  \multicolumn{2}{c |}{\multirow{2}{*}{\xmark}}&  \multicolumn{2}{c |}{\multirow{2}{*}{\cmark}}&\multirow{2}{*}{\xmark}\\ \hline
      Flashing light &  \multicolumn{2}{c |}{\multirow{2}{*}{\xmark}}&  \multicolumn{4}{c |}{\multirow{2}{*}{\cmark}}&  \multirow{2}{*}{\xmark}&  \multicolumn{2}{c |}{\multirow{2}{*}{\cmark}}\\ \hline
      Reflective objects & \multicolumn{3}{c |}{\multirow{2}{*}{\cmark}}& \multicolumn{2}{c |}{\multirow{2}{*}{\xmark}}& \multicolumn{2}{c |}{\multirow{2}{*}{\cmark}}& \multicolumn{2}{c |}{\multirow{2}{*}{\xmark}}\\ \hline
      Preferred objects & \multicolumn{3}{c |}{\multirow{2}{*}{\xmark}}  & \multicolumn{6}{c |}{\multirow{2}{*}{\cmark}}\\ \hline
    \end{tabular}

    \vspace{3mm}
    \caption{The nine scenes (image sequences) in our AE dataset. See Section~\ref{sec:dataset} for detail of the table and video/image sequence description.}
    \label{table:scene_settings}
    \end{table*}

\section{Platform}

\begin{figure}[ht]
	\centering
	\includegraphics[width=0.5\textwidth]{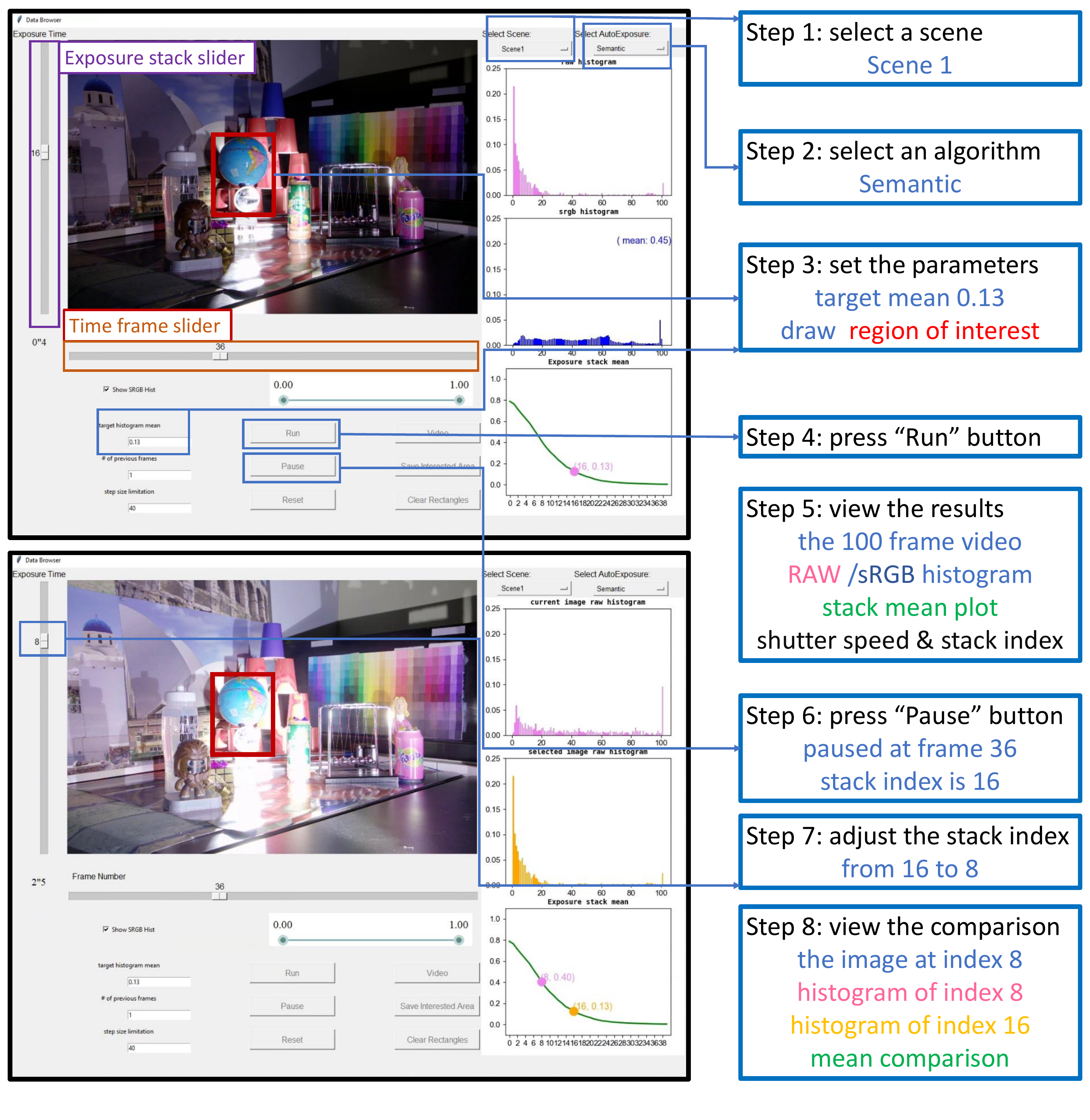}
	\caption{The basic steps for using our AE platform. The user selects a scene and an algorithm.  Parameters of the AE algorithm can adjust. After the AE algorithm runs, the platform plays the output images and the corresponding plots at 10 FPS. The user cause a ``pause'' at any time frame to adjust the exposure stack slider for comparison. Image histogram for each frame are also shown.}
	\label{fig_guil}
\end{figure}

A Python-based AE evaluation platform is also developed to work with our dataset. Figure~\ref{fig_guil} shows the platform's interface and the basic workflow of testing an AE algorithm. On the top right of the window, two drop-down menus are provided for the scene and algorithm selection. Users can adjust the parameters for the various AE algorithms described in Section~\ref{algorithms}.  The user can also select the starting shutter speed as the input to the AE algorithm.  After the parameters for an AE algorithm are set, the AE algorithm can be applied, where it will select a single exposure per time-step that can be played back in our GUI or saved to a video.

%Our platform also allows a visual exploration of the dataset and an AE solution. Sliders allow navigation of the exposure and temporal dimensions. For example, the upper-half image in Figure~\ref{fig_guil} shows the 16th image (shutter speed 2/5 s) of the 36th frame in scene 1 of our dataset. This image could either be selected manually by adjusting the sliders or have been selected by an AE algorithm as the optimal frame for this time step. The second part of the visualization shows three additional plots. These plots are displayed to help the user understand the AE algorithm's performance. In particular, the user can see a different combination of plots when operating the interface. An example is shown in Figure~\ref{fig_guil}. After taking the user's input in  ``Step 3,'' and running the ``semantic'' algorithm described in Section~\ref{algorithms:Histogram Manipulation}, the GUI displays the plots, which allows the user to view the RAW image histogram, the processed image(sRGB) histogram and the current frame stack mean values.

%The user may pause the algorithm at any frame and adjust the vertical slider (to visualize an image with another EV from the same frame); then, the plots will switch to the mode as in the bottom part of Figure~\ref{fig_guil}. In this case, the user sees the difference between the current image and the one selected by the AE algorithm in terms of the RAW histogram and mean.

\section{AE Algorithms}
\label{algorithms}
For our evaluation of AE algorithms, we implemented two content-agnostic and two semantic AE algorithms. The first AE was a content-agnostic global algorithm driving the mean to a target value~\cite{histogramMean}. The second content-agnostic AE focused on entropy maximization~\cite{entropy3}. Next, we implement a semantic AE that uses manually drawn bounding-boxes to give preference to certain regions. Finally, we implement our own semantic AE guided basd on a fast saliency method~\cite{zhang2015MBD}.

\begin{figure*}
\begin{center}
\includegraphics[width=0.98\textwidth]{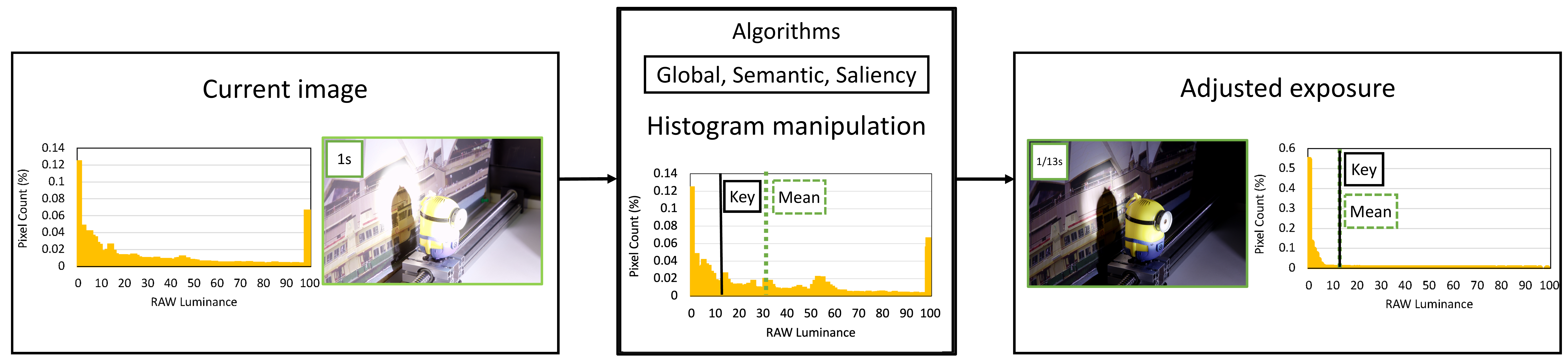}
\end{center}
\caption{The basic steps for most AE algorithms. First, each AE algorithm uses different criteria to decide how to construct a histogram from the current image frame.  Second, the mean value of the histogram is compared to a user-specified key value. The shutter speed is adjusted to be either longer or shorter such that the next image's mean histogram value will better match the specified key.}
\label{fig_alg}
\end{figure*}

Most AE algorithms (with the exception of the entropy methods) can be modeled with two steps shown in Figure~\ref{fig_alg}.
 \begin{enumerate}
  \item Histogram manipulation: each AE algorithm determines which pixels in the image contribute to the image histogram.
  \item Exposure modification: the shutter speed is adjusted such that the histogram's mean value is shifted towards a user-defined target value (referred to as the `key').
\end{enumerate}

Each of these steps is discussed in more detail in the following.

\subsection{Histogram manipulation}
\label{algorithms:Histogram Manipulation}
A histogram $H$ can be represented by a vector of RAW pixel values $H_p = {[p_1, p_2, ..., p_n]}$ and corresponding weights $H_w = {[w_1, w_2, ...., w_m]}$. Each corresponding weight can be considered how important a pixel is to the AE algorithm used. Typically, AE histogram manipulation consists of changing the histogram by modifying $H_w$.

We look closely at three types of AE algorithms that perform histogram manipulation. The {\it global} algorithm uses the full image histogram. The {\it semantic} algorithm only uses the histogram within a specified area. We introduce our own method: a {\it saliency} AE algorithm that uses a saliency map to build the histogram weights. An example of each algorithm is shown in Figure~\ref{fig_alg_breakdown}.

The {\it global} method weights all pixels equally. Our {\it semantic} method is similar to the {\it global} method, but uses only pixels within a defined bounding-box to construct the histogram.  In our work, we have manually drawn this bounding-box to emulate different detectors, such as a face detector or object tracking~\cite{face_auto_exposure}. Finally, we introduce a simple {\it saliency} method, where we first run a fast saliency~\cite{zhang2015MBD} detector on the image.  The saliency map is thresholded and pixels above the threshold contribute more to the final histogram (see supplemental materials for more details). If no pixel is labeled salient, the saliency method reduces to the global algorithm where all pixels have equal importance.

After using the algorithm-specific weighting function, all algorithms implemented a histogram clipping of saturated pixels. This was done by zeroing the weights of most (99\%) pixels with intensity greater than 0.9; we allow a small percentage (1\%) of these thresholded pixels to contribute to the histogram to avoid an empty histogram. 

\subsection{Exposure modification}
% Sai/Beixuan: Read this part carefully. 
%Response (Sai) - Looks good to me
The mean of a histogram $H$ is given by a weighted average where $H_w$ is normalized to add to 1. The camera-specific key is typically a RAW value that maps to half-brightness (0.5) in sRGB after being passed through the camera-specific image signal processor (ISP). A typical RAW key value would be between 0.18 and 0.23 because the sRGB gamma maps these values close to 0.5 in sRGB. However, we used a key value of 0.13 because this gives a result close to half-brightness in sRGB due to the additional tone manipulation in the camera pipeline~\cite{Abdelhamed2018} used to process the RAW image. 

% beixuan: 0.5^2.2 = 0.21, 0.5^2.4 = 0.19. I am not sure if it is accurate to say 0.18 is the gamma maped to 0.5. And also I think the tone

The goal of exposure modification is to bring the mean of a histogram to the camera-specific key. Since exposure has a linear relationship to the RAW image, the exposure modification can be calculated as a scale between the key and the current mean of the histogram. In our implementation, each algorithm's modified histogram is driven to the key by adjusting the shutter speed up or down.

\subsection{Entropy AE}

\begin{figure*}[h]
  \begin{center}
  \includegraphics[width=1\textwidth]{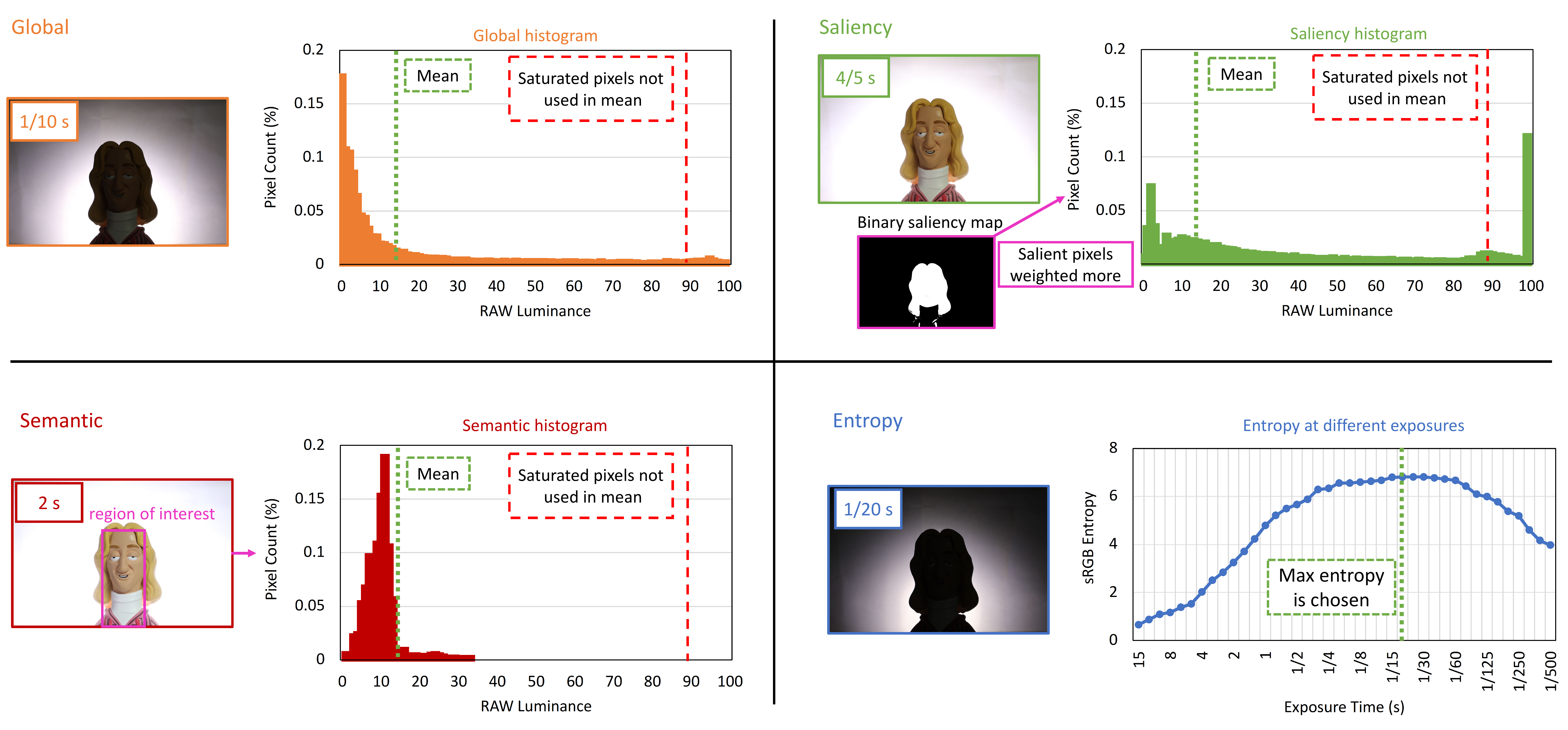}
  \end{center}
  \caption{An overview of the four AE algorithms used in this work. A global AE (top left) that uses the whole image to build the histogram.  A saliency AE (top right) that weights salient pixels with higher weight. A semantic AE (bottom left) that uses the bounding box we provide to build the histogram. Finally, an entropy AE (bottom right) that chooses the exposure with maximal entropy. Here, we show the images chosen by all of these algorithms for scene 4 at time step 59.}
  \label{fig_alg_breakdown}
  \end{figure*}

We also implemented the {\it entropy} AE algorithm by Zhang et al.~\cite{entropy3}. Since this method operates on a post-processed image, we perform our entropy calculation in the post-processed sRGB space. For each time step, we compute the exposure that maximizes the entropy across the exposure stack. An example of this entropy maximization is shown in Figure~\ref{fig_alg_breakdown}.  At each time step the maximum entropy results in an ``optimal'' sequence of exposure indices. This is possible because our dataset gives us access to all exposures for any time step. Note that in practice, a real AE system would not have access to all the exposure results at a given time step, so this method has an unfair advantage.

%caption needs work about bounding box -

\subsection{Reduced-size AE}
\begin{figure}[h]
  \begin{center}
  \includegraphics[width=0.48\textwidth]{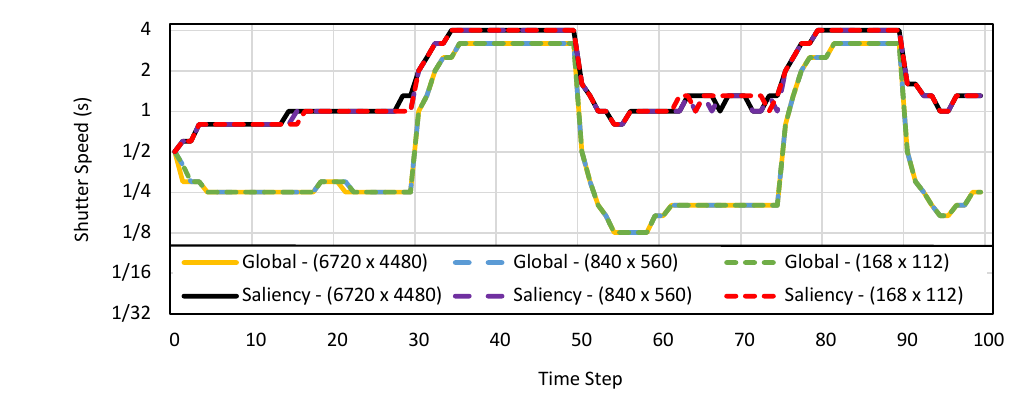}
  \end{center}
  \caption{A comparison of global and saliency AE run at three different scales ($6720 \times 4480$ [full size], $840\times560$, and $168\times112$) on scene 6. There are minimal differences in AE algorithm output.}
  \label{fig_different_scale}
  \end{figure}

%Not sure if this belongs here or just at the top of experiments section - What do you think?
Most camera AE systems subsample the RAW image. Subsampling maintains the shape of the histogram and allows for faster AE operation. We verified scale doesn't affect AE output drastically by applying all four AE algorithms at three different scales (full size, 840 $\times$ 560, 168 $\times$ 112) across all scenes. Reducing the image size to 840 $\times$ 560 resulted in an average EV change of 0.076 across all frames, scenes, and AE algorithms (i.e., minimal impact); the average EV change was 0.081 when reducing to an image size of 168 $\times$ 112. Figure~\ref{fig_different_scale} shows an example of how exposure sequences selected by global and saliency algorithms run at different scales differ marginally. 

%caption needs work about bounding box -

\section{Experiments}
\label{sec:experiments}

One benefit of our dataset and platform is the ability to evaluate different AE algorithms on the same input.  Since each algorithm has different criteria about which image pixels or pixel intensities should contribute to the determination of the exposure, we performed a user study to see if users have a preference. 

\subsection{User study}
We conducted a forced-choice user study to compare all four AE algorithms on the nine scenes from our dataset. We evaluated the algorithms by compiling overall user preference scores for each algorithm.

To perform our study, we generate simulated videos for all nine scenes in our dataset with four AE algorithms (global, semantic, saliency, and entropy). Videos using semantic AE used the bounding boxes provided with the dataset for all frames. For saliency AE, the first time step assumes no salient pixels. All other time steps used the saliency map generated from the previous frame's post-processed sRGB image.  Settings for the different algorithms and examples of generated videos are provided in the supplemental materials. Figure~\ref{fig_videoframes} shows an example output for the different methods on a particular scene. 

%Michael - Is the first time step unneccesary?
% MSB: I don't think it is necessary, we can move it to the supp materials
\begin{figure*}
\begin{center}
\includegraphics[width=1\textwidth]{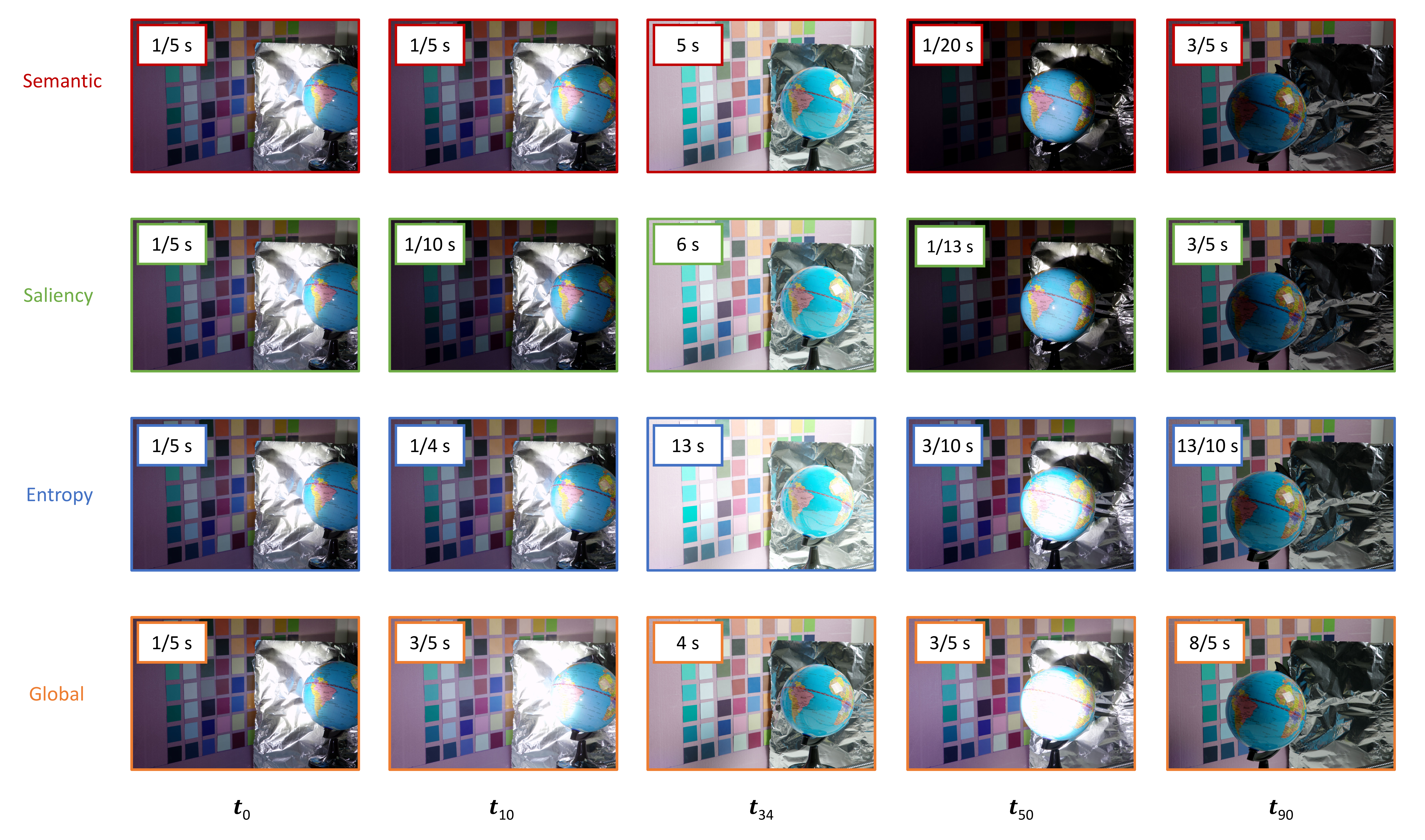}
\end{center}
\caption{The frames chosen by different AE algorithms for scene 7 at five time steps. All AE algorithms started at the same exposure.}
\label{fig_videoframes}
\end{figure*}

For each time-frame, we calculate the optimal index that brings the algorithm's modified histogram mean closest to the key. Then, we smooth the indices using a history of four time-steps to have smooth transitions as the exposure changes; this was inspired by Aboulaim et al.~\cite{Abuolaim2018}, who showed users preferred videos with smooth transitions over those with abrupt changes.

Our study involved 33 people. The age of participants ranged between 22 and 39 (mean = 24.4, standard deviation = 3.3). The cohort included 22 males and 11 females.  Participants performed the study in a dark room on a 15" MacBook Pro.  
Participants were asked to sit centered in front of the computer screen and read the instructions. Then, participants did 54 forced-choice comparison trials; this took 10-15 minutes per participant. In addition, the trial order was randomized for each participant.

For each trial, a participant would view two videos rendered with different AE algorithms synchronously. Videos were 10 seconds long (10 FPS) and looped until the user selected a preferred video---that is, forced choice. Participants selected their preferred video by using the left and right arrows on the keyboard. The option to select a video was enabled only after watching the video fully one time.

This study utilized a 4$\times$9 within-subjects design with the following independent variables and levels:  AE algorithm: semantic, saliency, entropy, global; Scene: 1,2, ..., 9.

The dependent variable measured was user preference. The average number of votes for an AE algorithm is bounded between 0 and 3 because each algorithm is only compared in 3 trials per scene. User preference normalizes the average number of votes by 3, so the metric is bounded between 0 and 1.

Each scene requires six trials to have all pairings between the four AE algorithms. Thus, the total number of trials was 33 participants $\times$ 9 scenes $\times$ 6 trials = 1782.

\subsection{User study results}

\begin{figure}
  \begin{center} \includegraphics[width=0.5\textwidth]{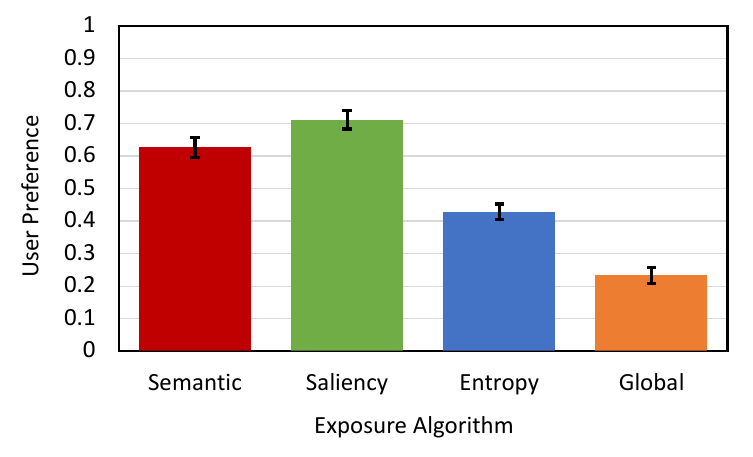}
  \end{center}
  \caption{The average preference of AE algorithms across all scenes. Error bars represent 95\% confidence interval bars. }
  \label{fig_average}
  \end{figure}

The average preference of all algorithms across the scenes was 0.71 for saliency, 0.63 for semantic, 0.43 for entropy, and 0.23 for global. Figure~\ref{fig_average} shows a comparison between the methods with 95\% confidence interval bars.

\begin{figure*}
  \begin{center}
  \includegraphics[width=1.0\textwidth]{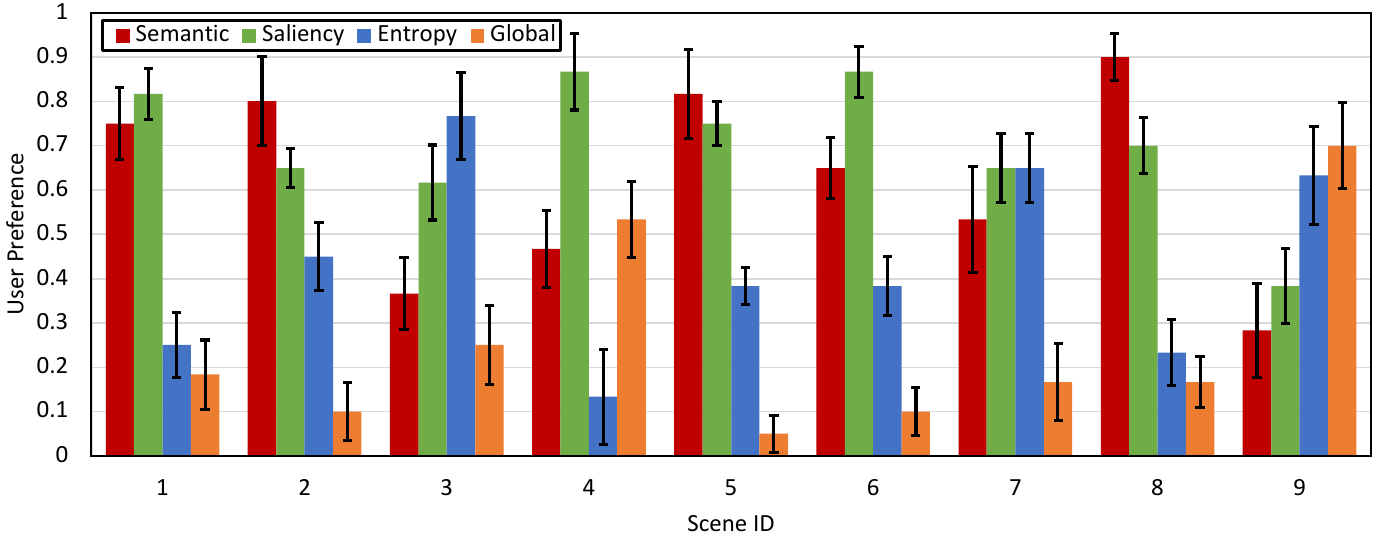}
  \end{center}
  \caption{The average preference of AE algorithms shown per scene. Error bars represent 95\% confidence interval bars. }
  \label{fig_scenebyscene}
  \end{figure*}

\begin{figure*}
\begin{center}
\includegraphics[width=1.0\textwidth]{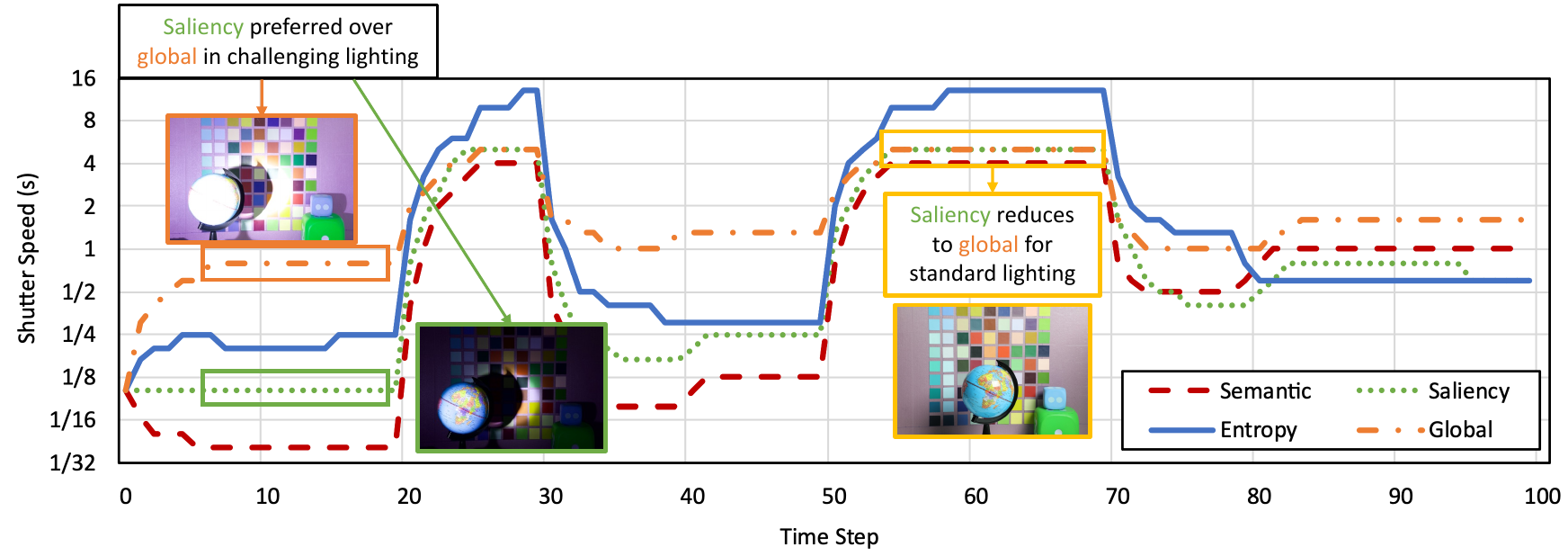}
\end{center}
\caption{A time plot showing how AE algorithms choose different exposures at various time steps. Here we show the results from scene 5. In challenging lighting conditions, saliency AE chooses an exposure that prioritizes the object at the expense of the background. However, in standard lighting, saliency and global AE choose the same exposure.}
\label{fig_timeplotexplain}
\end{figure*}

% Sai/Beixuan: Should be provide any citations for ANOVA ,Fisher LSD, and Bonferonni-Donn? - Response: these are old methods that originate in the early 1900s so there is no "original" citation, but I just cited a paper that discusses all three methods
An ANOVA~\cite{pereira2015overview} analysis was conducted and showed that the effect of the algorithm on preference was statistically significant $(F_{8,72} = 103.144, p < .0001)$. Additionally, the post-hoc Fisher LSD and Bonferonni-Donn comparison tests~\cite{pereira2015overview} resulted in a significant difference between all pairs of algorithms.

In addition, Figure~\ref{fig_scenebyscene} shows a breakdown of the preference of algorithms per scene. We see that the saliency AE performed well across all scenes except scene 9. Most of the time, we found that users were willing to have over/under-exposed backgrounds if it meant a foreground object (often moving) was well exposed. However, for scene 9, users preferred a result where the foreground object (a figurine) was dimly lit so they could see the background; thus, saliency and semantic AE performed poorly. 
%% Sai/Beixuan, we have never mentioned gamma before have we? %Response - Removed it

There was a clear preference for the saliency and semantic algorithms, because they prioritized the objects participants were more interested in viewing. Objects of interest typically are moving, familiar (e.g., face), or centered in the frame. To better understand why saliency AE had higher user preference than other methods, we generated a plot of exposures selected by different AE algorithms on different scenes; Figure~\ref{fig_timeplotexplain} shows an example for scene 5. We noticed that the global and saliency method chose similar exposures at time steps with less challenging lighting. However, for time steps with challenging conditions (e.g. bright globe), saliency AE chooses an exposure with more poorly exposed pixels that keeps the preferred object at a proper exposure. 

Even though our implementation of entropy AE was advantaged, it did not perform well. This might indicate entropy is not a good metric when designing AE algorithms for human viewing.

\section{Concluding remarks}

This paper captured a 4D dataset for studying AE algorithms in challenging lighting environments. In particular, we used a stop-motion setup to capture a temporal sequence with an exposure stack at each time step. We also developed a software platform to test different AE algorithms and visualize the algorithm's solution with respect to the full solution space. Our overall dataset consists of 36,000 images that emulate nine scenes, each with 100-time steps and each time step with 40 exposures. Our scenes include a variety of objects, object motion, and lighting configurations.
%ADDED HERE->
We implemented four AE algorithms (global, semantic, saliency, and entropy) and tested them on all scenes in our dataset. We used our platform to produce videos from these algorithms' output to conduct a user study to determine preference between methods. We found that users preferred semantic and saliency methods, where a region of interest was weighted more in the exposure decision. For time steps with relatively standard lighting conditions (e.g., Figure~\ref{fig_timeplotexplain} between time steps 55-65), there was no significant difference between AE algorithms. Our dataset and code is on the project website: \url{https://ae-video.github.io}.

\paragraph{Acknowledgments}~Thanks to Rajat Ajayakumar for assisting in data capture. This study was funded in part by the Canada First Research
Excellence Fund for the Vision: Science to Applications (VISTA) programme and an NSERC Discovery Grant.

{\small
\bibliographystyle{ieee_fullname}
\bibliography{egbib}
}

%Fisher LSD and Bonferonni-Donn show differences between all pairs
%
%Summary statement:
%The effect of algorithm on votes was statistically significant (F(4, 108) = 103.144, p < .0001).

%Two Way
%C(scene)            7.416923e-27     8.0  2.466494e-27   1.000000e+00
%C(method)           9.003475e+02     4.0  5.988203e+02  2.192109e-304
%C(scene):C(method)  4.019798e+02    32.0  3.341955e+01  3.970058e-149

\onecolumn

\begin{center}
      {\Large \bf Supplementary Material: Examining Autoexposure for Challenging Scenes \par}
      % additional two empty lines at the end of the title
      \vspace*{24pt}
      {
      \large
      \lineskip .5em
      \par
      }
      % additional small space at the end of the author name
      \vskip .5em
      % additional empty line at the end of the title block
      \vspace*{12pt}
   \end{center}
   \begin{multicols}{2}

\setcounter{section}{0}

Our supplementary material includes additional AE algorithm implementation details and dataset visualizations.  

%%%%%%%%% BODY TEXT
\section{AE algorithms}

The main paper discussed four AE algorithms. Here we discuss the settings for these algorithms. In particular, we provide details on the global, semantic, and saliency AE algorithms. These methods use the same AE mechanism of adjusting the shutter speed based on histogram means and its relationship to a target value (key). These methods differ in how they construct (or manipulate) the image histogram. This is followed by a discussion on the entropy AE method that uses a different strategy to decide the optimal AE value.

\noindent{\textbf{Histogram manipulation AE}}~Recall that a histogram $H$ is represented by a combination of its pixel values $H_p$ and corresponding weights $H_w$.  

\begin{figure*}[h]
	\begin{center}
	\includegraphics[width=1\textwidth]{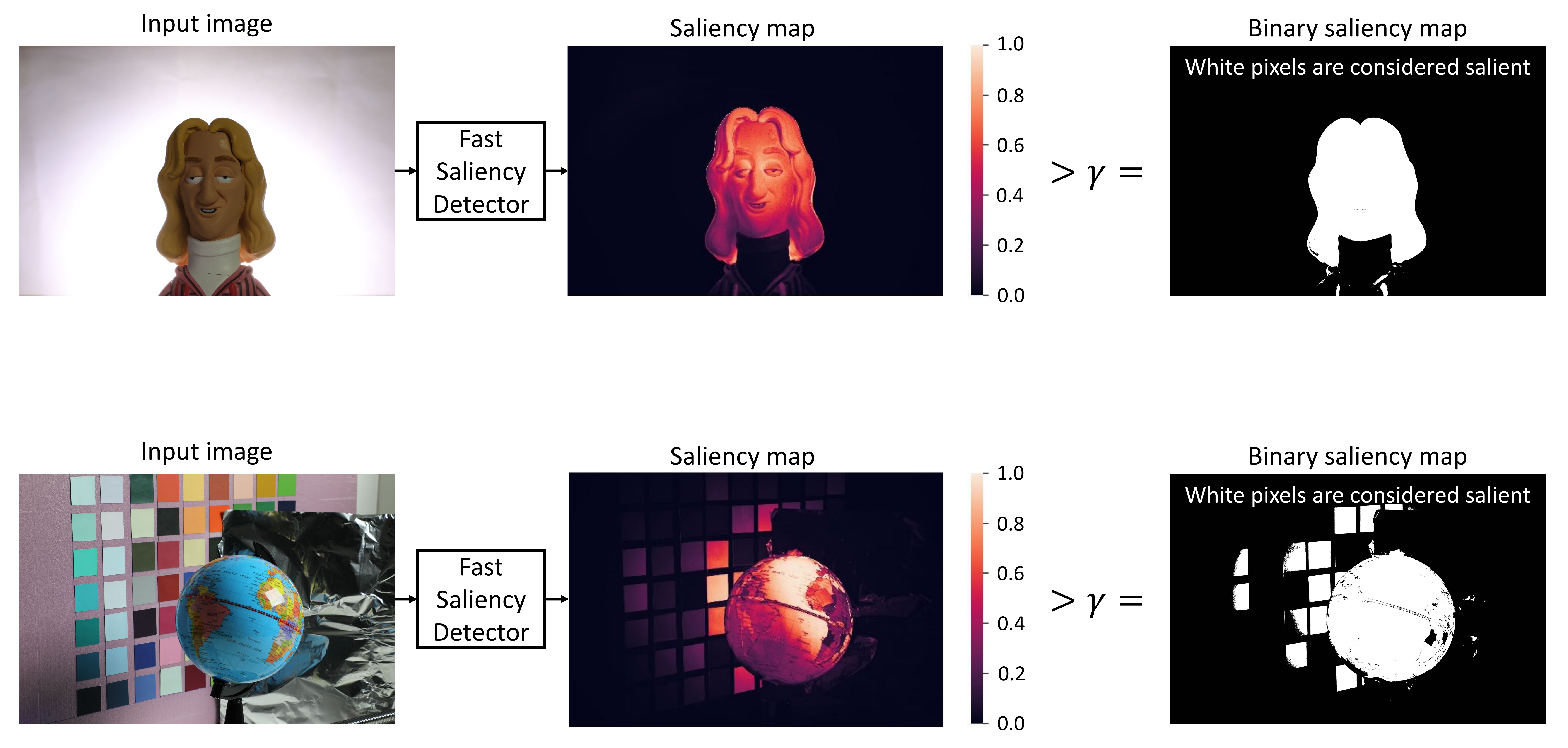}
	\end{center}
	\caption{Example binary saliency maps generated from two images in our dataset. Each binary saliency map is generated by a fast saliency method~\cite{zhang2015MBD} thresholded with $\gamma = 0.1$. }
	\label{fig_saliency}
\end{figure*}

\begin{itemize}
    \item Global AE gives all pixels within the image an equal weight. 
    \item Semantic AE uses the hand-drawn bounding boxes provided for each frame within a scene. All pixels within the bounding boxes are given a weight of 1; pixels outside the bounding boxes are given a weight of 0. 
    \item Saliency AE uses the last time step's saliency map to generate weights for the current time step's histogram. To generate the salient map, we use the fast saliency detector~\cite{zhang2015MBD} on the sRGB image produced by the prior time step. This produces a saliency map where each pixel has an associated value between 0 and 1. Next, we build a binary saliency map where any pixels above $\gamma$ are considered salient; $\gamma$ can be interpreted as the ``sensitivity'' of the saliency detector. Salient pixels were given a weight of $\beta$, and the remaining pixels were given a weight of $1$. For our implementation, we set $\gamma=0.1$ and $\beta=14$.  Figure~\ref{fig_saliency} contains two examples of our generated binary saliency maps. Additionally, for the first time step in a scene, we consider no pixels to be salient. 

\end{itemize}

After using the algorithm-specific weighting function, all algorithms implemented a histogram clipping of saturated pixels. This was done by zeroing the weights of all pixels with an intensity greater than 90\% of the maximum intensity value in the image.  

\noindent{\textbf{Entropy AE}}~For entropy AE, the frame with the maximum entropy within sRGB space is chosen for each time step. Our implementation of entropy AE is advantaged because a typical implementation has to perform a local search to find the maximum entropy image~\cite{entropy3, entropy1, entropy2}, however, in our case, we search the entire exposure stack to find the optimal entropy image.

\section{Platform GUI}

\begin{figure*}
	\centering
	\includegraphics[width=1\textwidth]{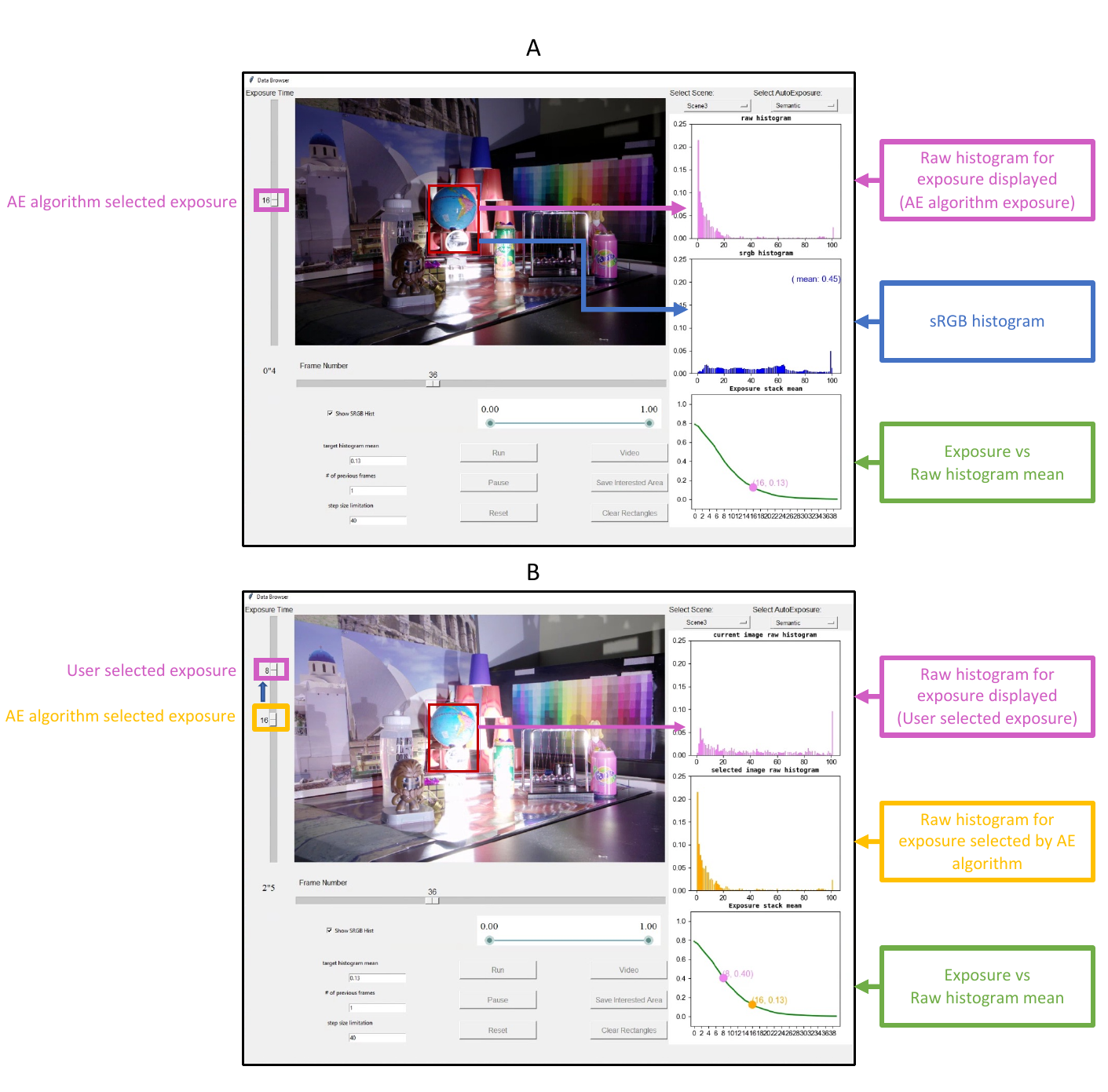}
	\caption{A visualization of our AE platform. When an AE algorithm runs (A), the platform displays the output images and the corresponding plots including ``RAW histogram'', ``sRGB histogram'' and ``Exposure vs RAW histogram mean''. The user is able to ``pause'' at any time frame to adjust the exposure stack slider for comparison (B). The plots of ``user selected RAW histogram'' and ``AE algorithm RAW histogram'' as well as their ``RAW histogram mean'' are also shown for contrast.}
	\label{fig_guil1}
\end{figure*}

In addition to the basic function introduced in the paper, our platform allows visual exploration of the dataset and AE algorithms. Sliders allow navigation of the exposure and temporal dimensions. For example, Figure~\ref{fig_guil1}A shows the 16th image (shutter speed 2/5 s) of the 36th frame in scene 1 of our dataset. This image can be selected manually with the sliders or by an AE algorithm. Our platform displays three plots that help the user understand the AE algorithm's performance: the RAW image histogram, the processed image (sRGB) histogram, and the current frame stack mean values. The user may pause the algorithm at any frame and adjust the vertical slider (to visualize an image with another EV from the same frame); in this mode, the user is shown the difference between the current image and the one selected by the AE algorithm in terms of the RAW histogram and mean (visualized in Figure~\ref{fig_guil1}B).

\section{User study videos}
\begin{figure*}
	\begin{center}
 	\includegraphics[width=1\textwidth]{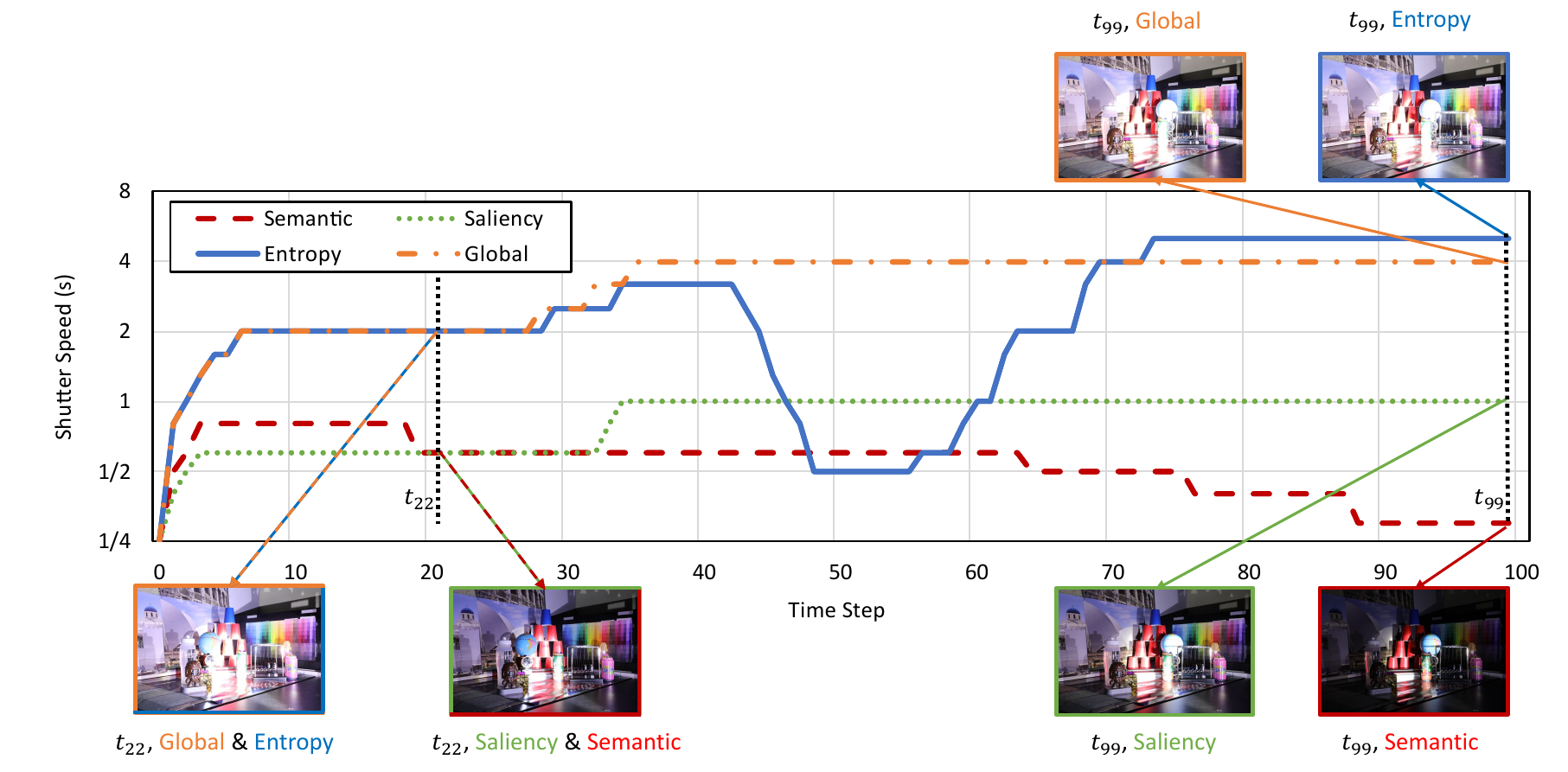}
	\end{center}
	\caption{Scene 1 time plot visualizing how AE algorithms choose different exposures at various time steps. This scene contains a moving mirror (high reflective object). Additionally, we visualize the images chosen by the AE algorithms at time steps 22 and 99.}
	\label{fig_timep1}
\end{figure*}

\begin{figure*}
	\begin{center}
	\includegraphics[width=1\textwidth]{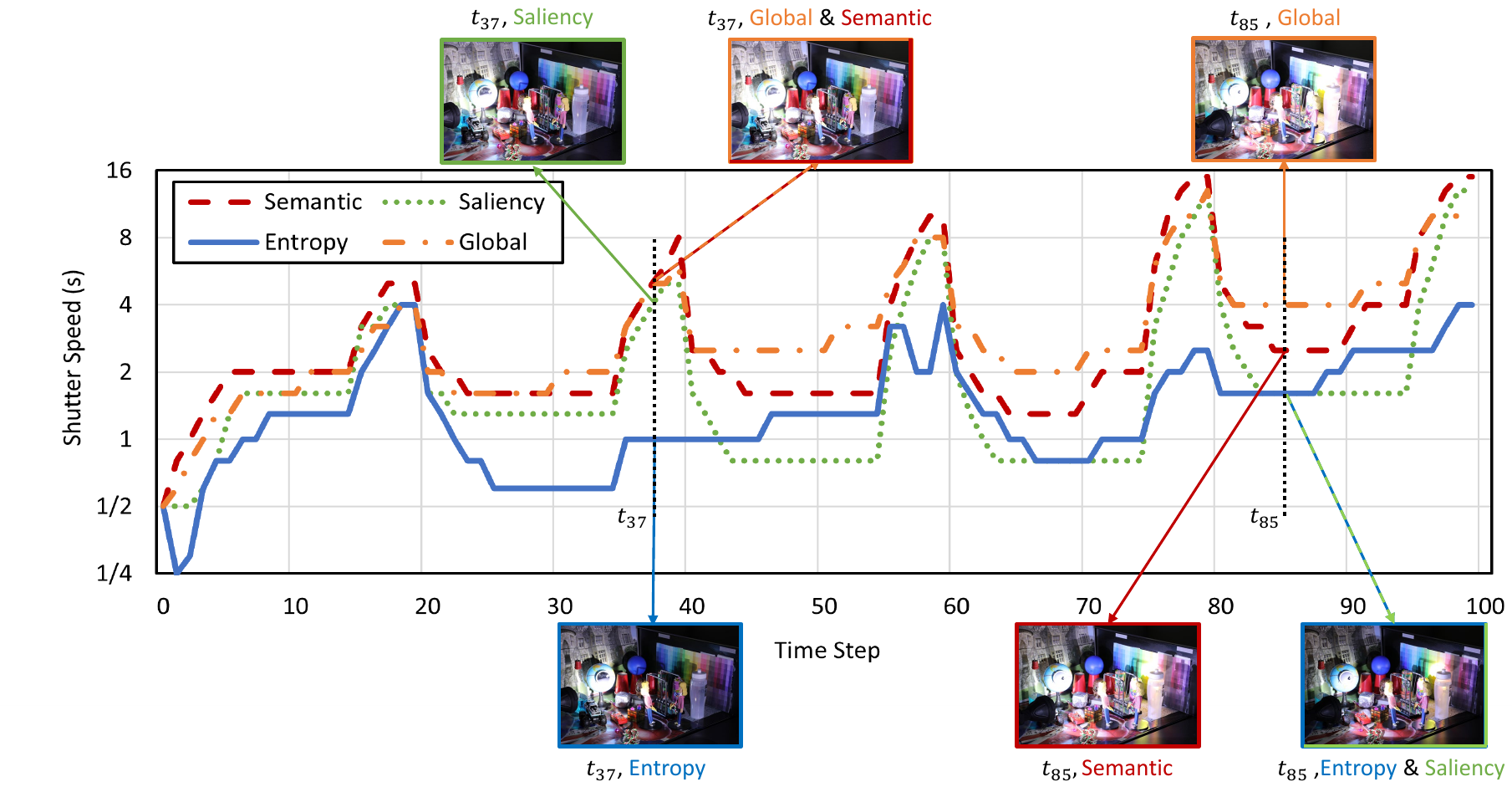}
	\end{center}
	\caption{Scene 3 time plot visualizing how AE algorithms choose different exposures at various time steps. This scene contains a moving, flashing light and a mirror (high reflective object). Additionally, we visualize the images chosen by the AE algorithms at time steps 37 and 85.}
	\label{fig_timep3}
\end{figure*}

\begin{figure*}
	\begin{center}
	\includegraphics[width=1\textwidth]{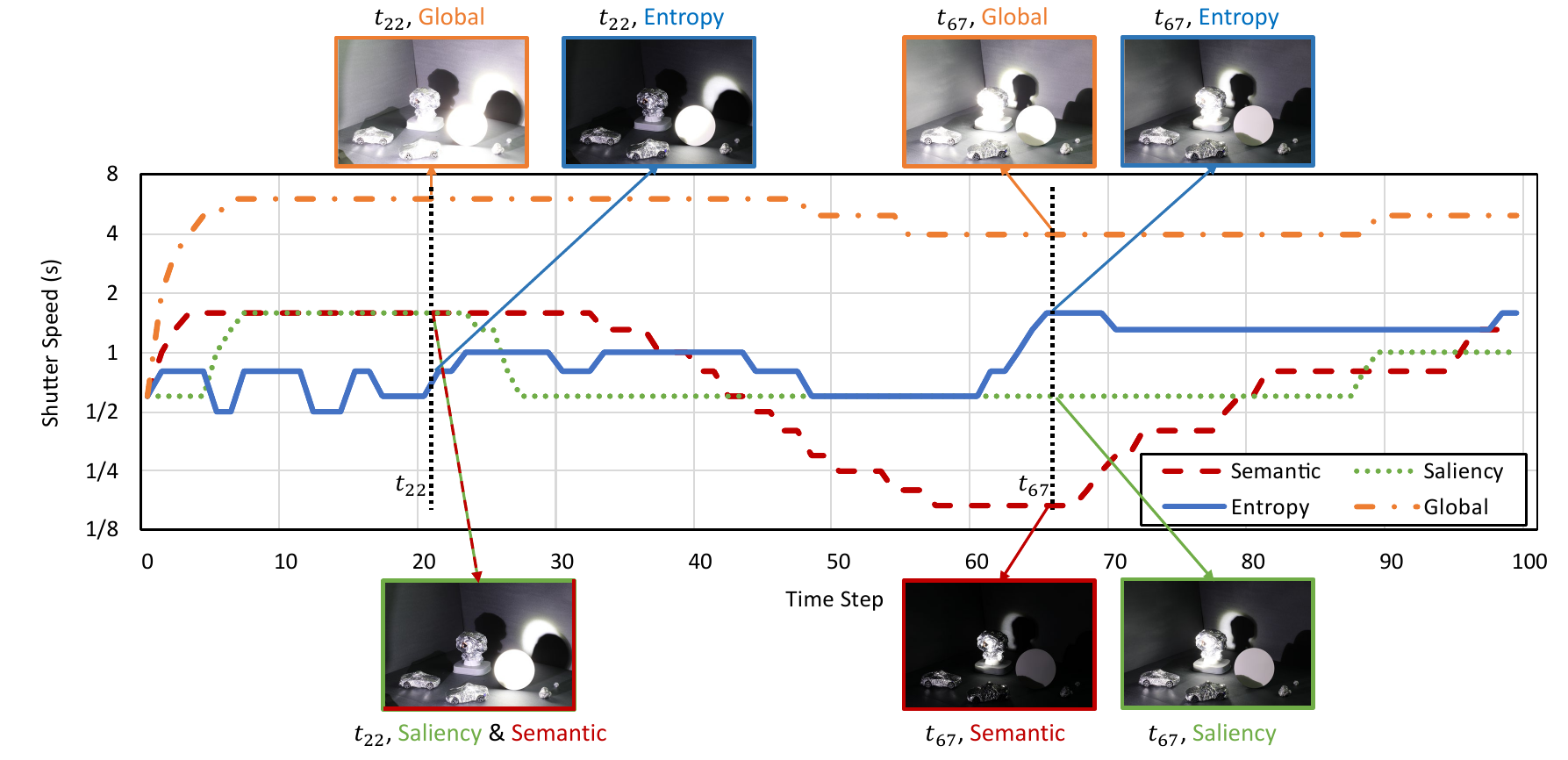}
	\end{center}
	\caption{Scene 7 time plot visualizing how AE algorithms choose different exposures at various time steps. This scene contains a moving high-lumen light and several highly reflective objects (doll, ball, and cars). Additionally, we visualize the images chosen by the AE algorithms at time steps 22 and 67.}
	\label{fig_timep7}
\end{figure*}

Videos produced by our platform GUI for all 4 AE algorithms on different scenes are available on our project page. Additionally, Figures~\ref{fig_timep1}-\ref{fig_timep7} show time plots comparing the AE algorithms on scenes 1, 3, and 7. 

\section{Dataset}
\begin{figure*}[ht]
	\begin{center}
	\includegraphics[width=1.0\textwidth]{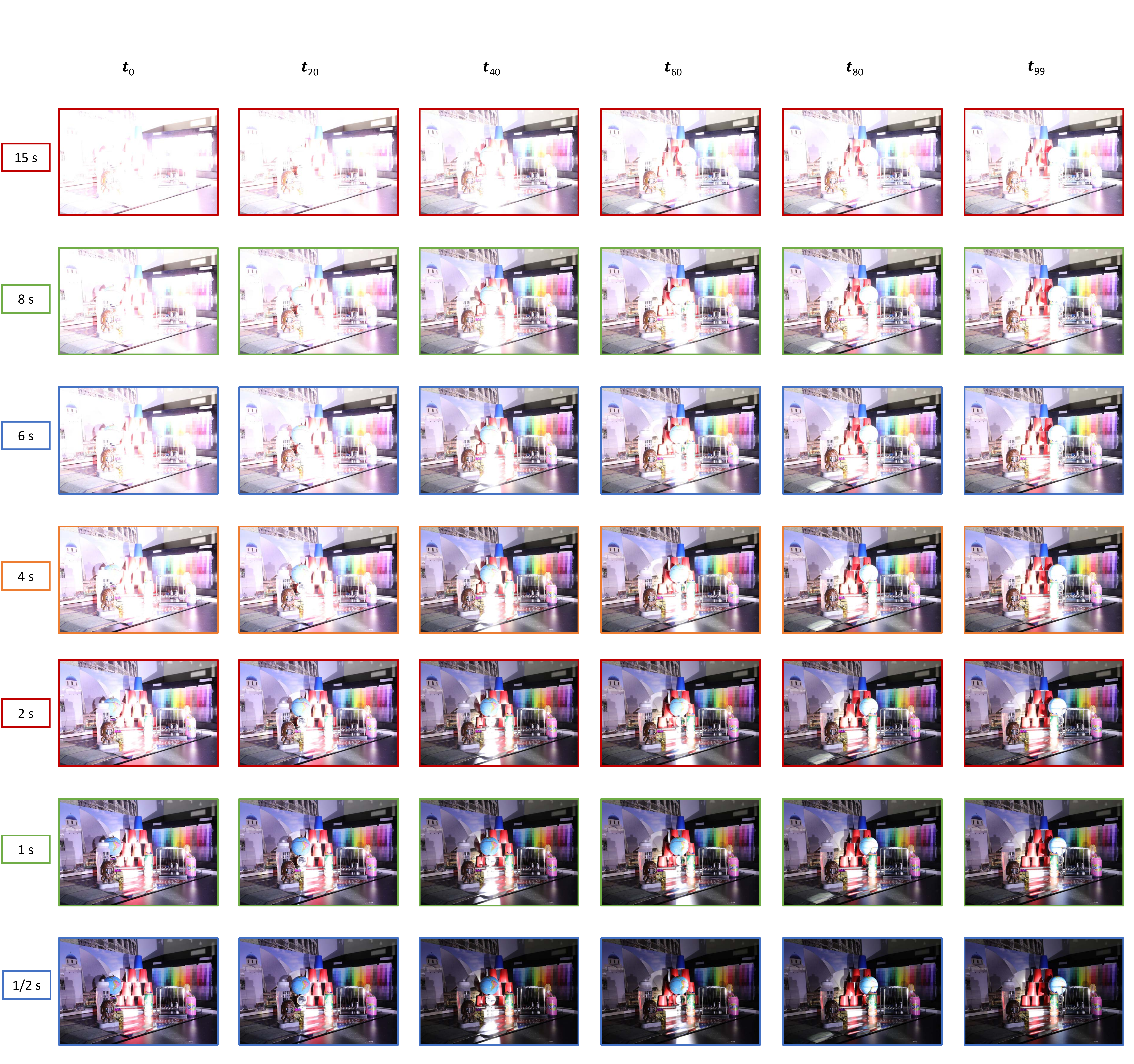}
	\end{center}
	\caption{Scene 1 exposure stack (15s - \(\frac{1}{2}\)s) at 6 different time steps.}
	\label{fig_data1_1}
\end{figure*}

\begin{figure*}
	\begin{center}
	\includegraphics[width=1\textwidth]{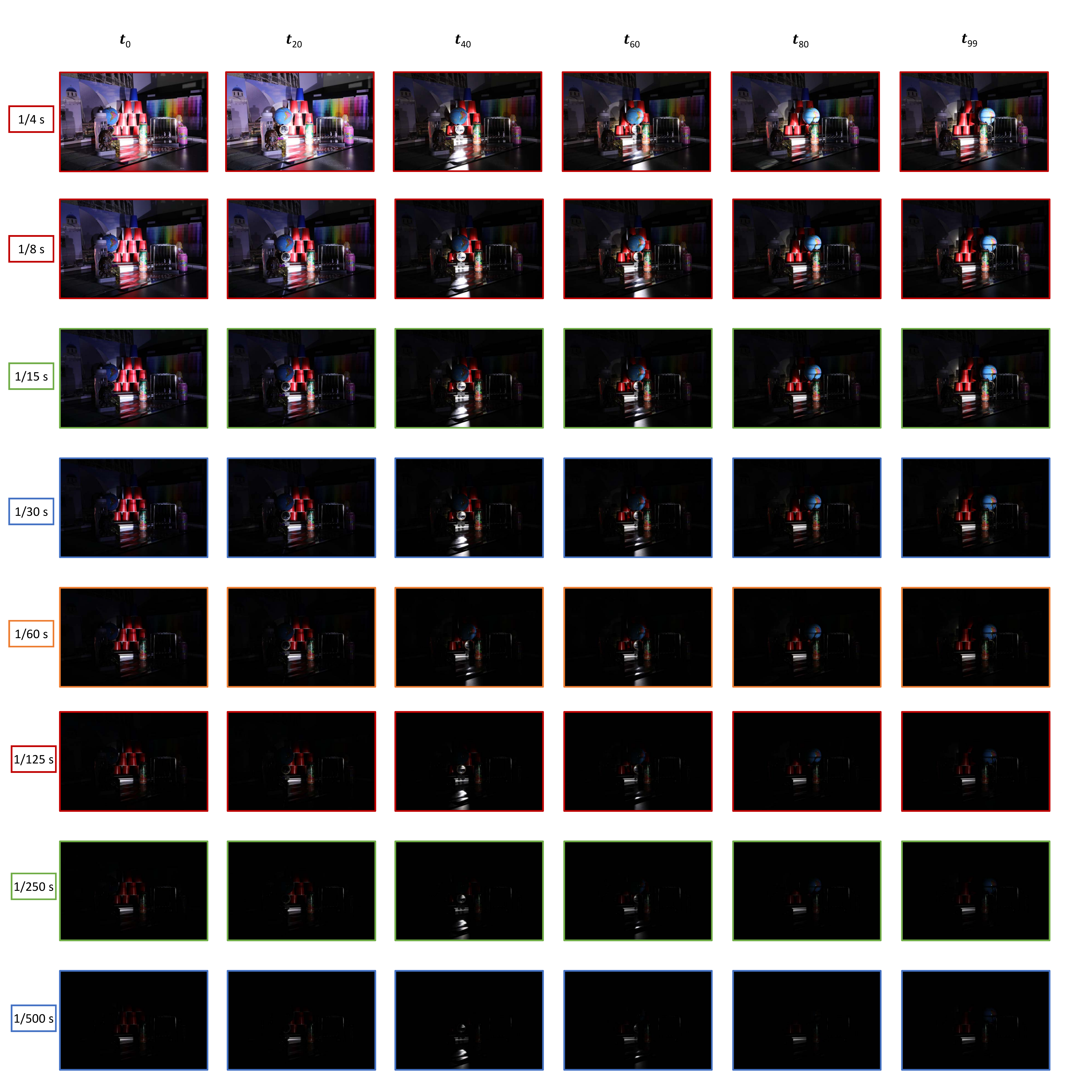}
	\end{center}
	\caption{Scene 1 exposure stack (\(\frac{1}{4}\)s - \(\frac{1}{500}\)s) at 6 different time steps.}
	\label{fig_data1_2}
\end{figure*}

\begin{figure*}
	\begin{center}
	\includegraphics[width=1\textwidth]{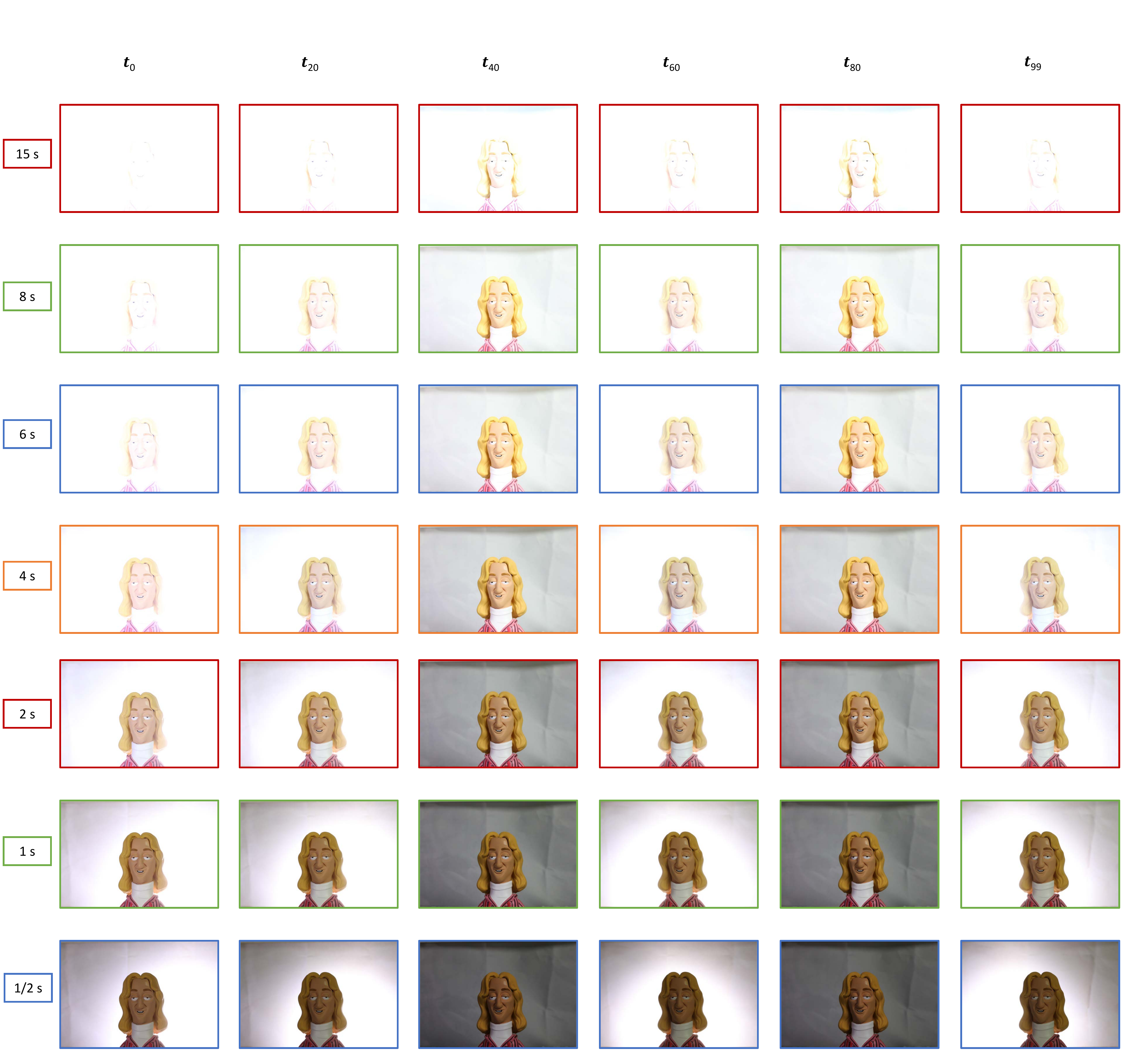}
	\end{center}
	\caption{Scene 4 exposure stack (15s - \(\frac{1}{2}\)s) at 6 different time steps.}
	\label{fig_data4_1}
\end{figure*}

\begin{figure*}
	\begin{center}
	\includegraphics[width=1\textwidth]{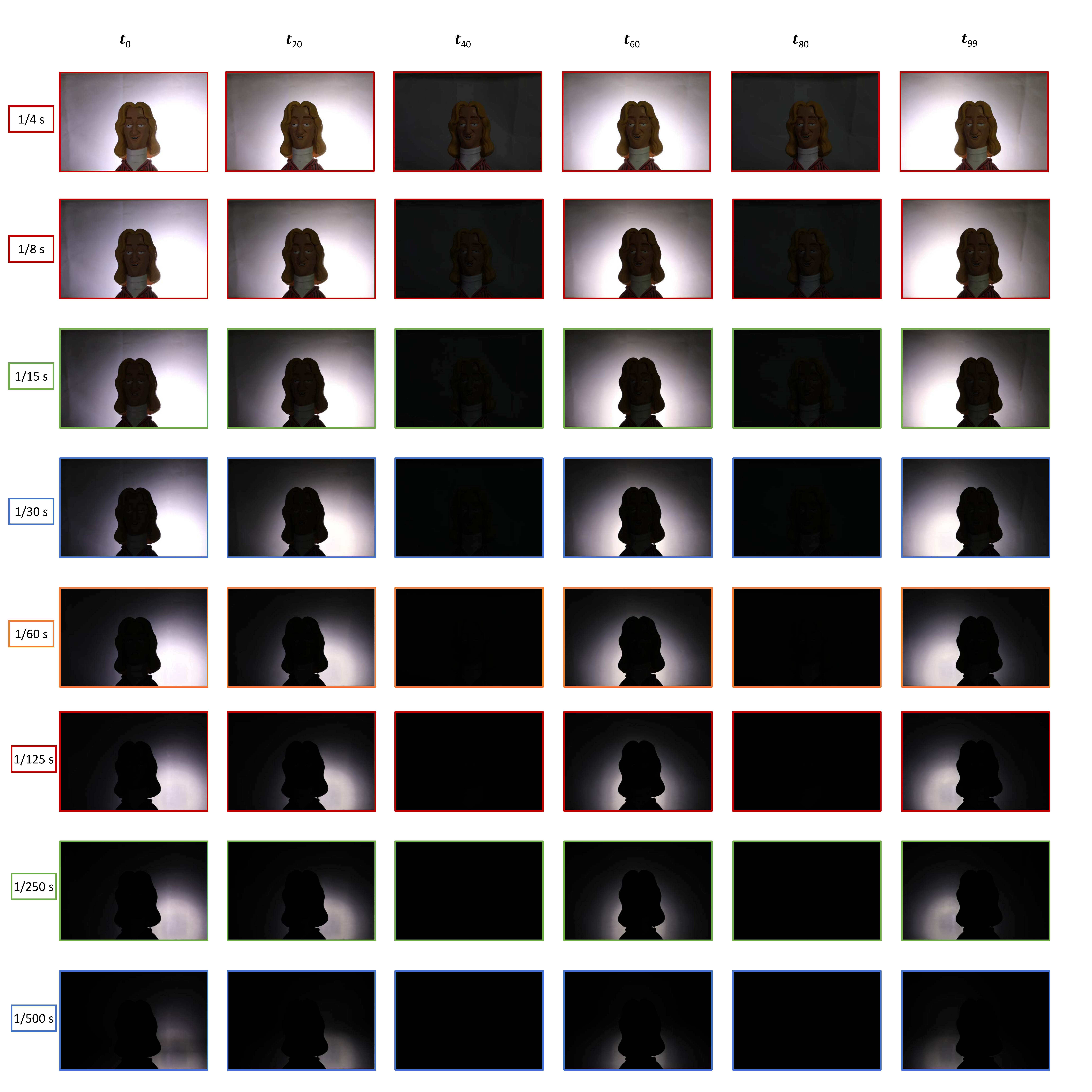}
	\end{center}
	\caption{Scene 4 exposure stack (\(\frac{1}{4}\)s - \(\frac{1}{500}\)s) at 6 different time steps.}
	\label{fig_data4_2}
\end{figure*}

\begin{figure*}
	\begin{center}
	\includegraphics[width=1\textwidth]{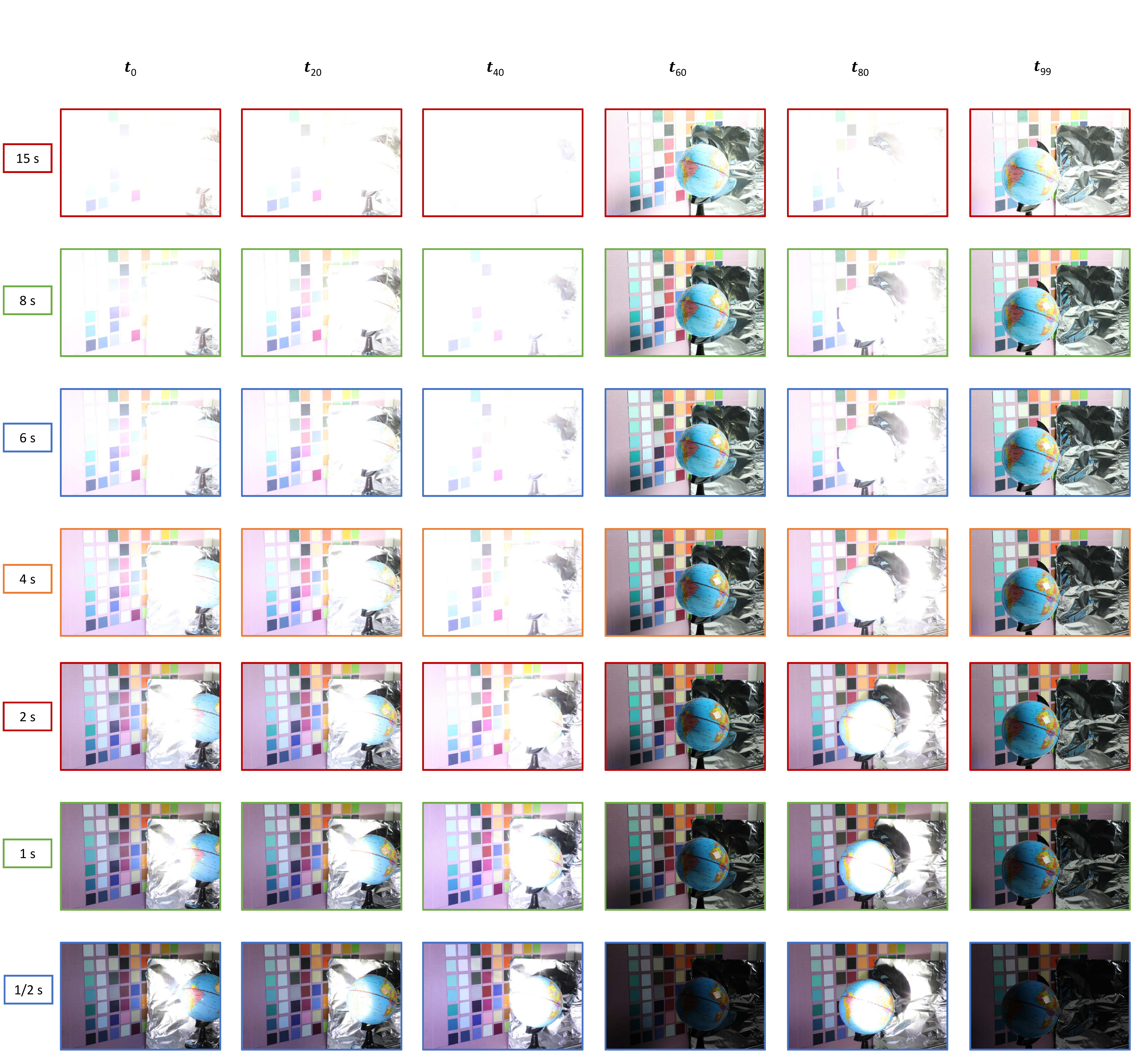}
	\end{center}
	\caption{Scene 6 exposure stack (15s - \(\frac{1}{2}\)s) at 6 different time steps.}
	\label{fig_data6_1}
\end{figure*}

\begin{figure*}
	\begin{center}
	\includegraphics[width=1\textwidth]{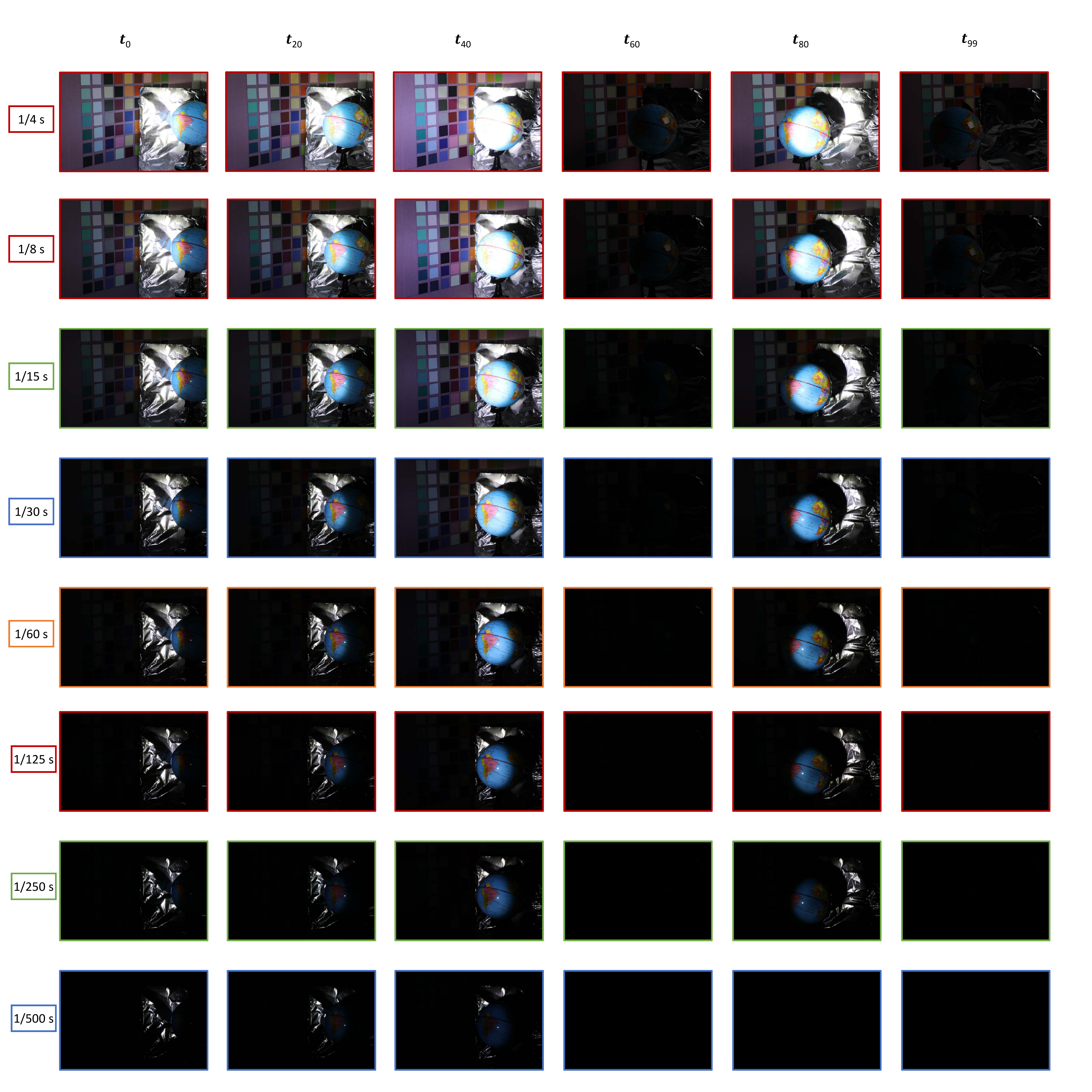}
	\end{center}
	\caption{Scene 6 exposure stack (\(\frac{1}{4}\)s - \(\frac{1}{500}\)s) at 6 different time steps.}
	\label{fig_data6_2}
\end{figure*}

\begin{figure*}
	\begin{center}
	\includegraphics[width=1\textwidth]{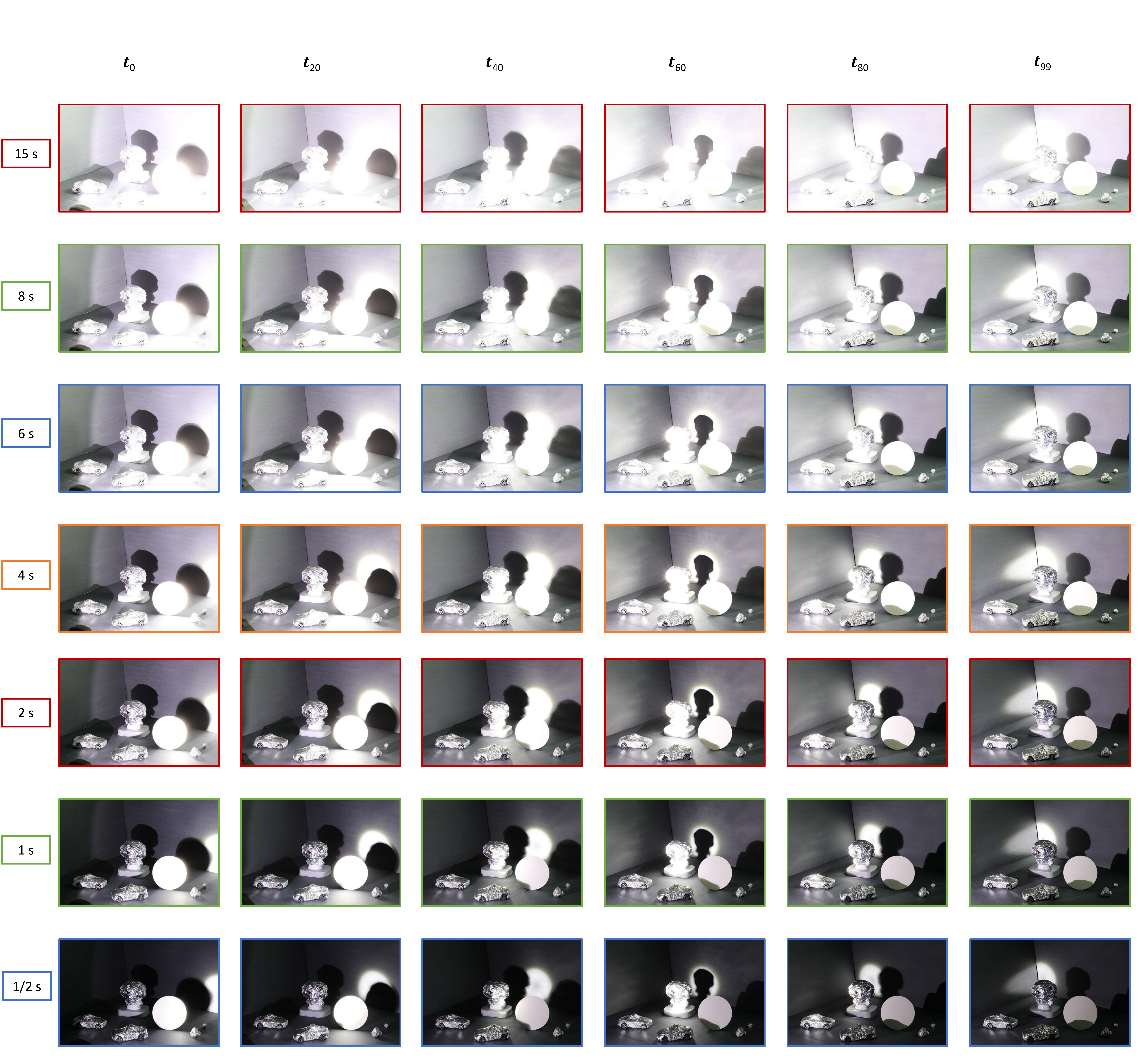}
	\end{center}
	\caption{Scene 7 exposure stack (15s - \(\frac{1}{2}\)s) at 6 different time steps.}
	\label{fig_data7_1}
\end{figure*}

\begin{figure*}
	\begin{center}
	\includegraphics[width=1\textwidth]{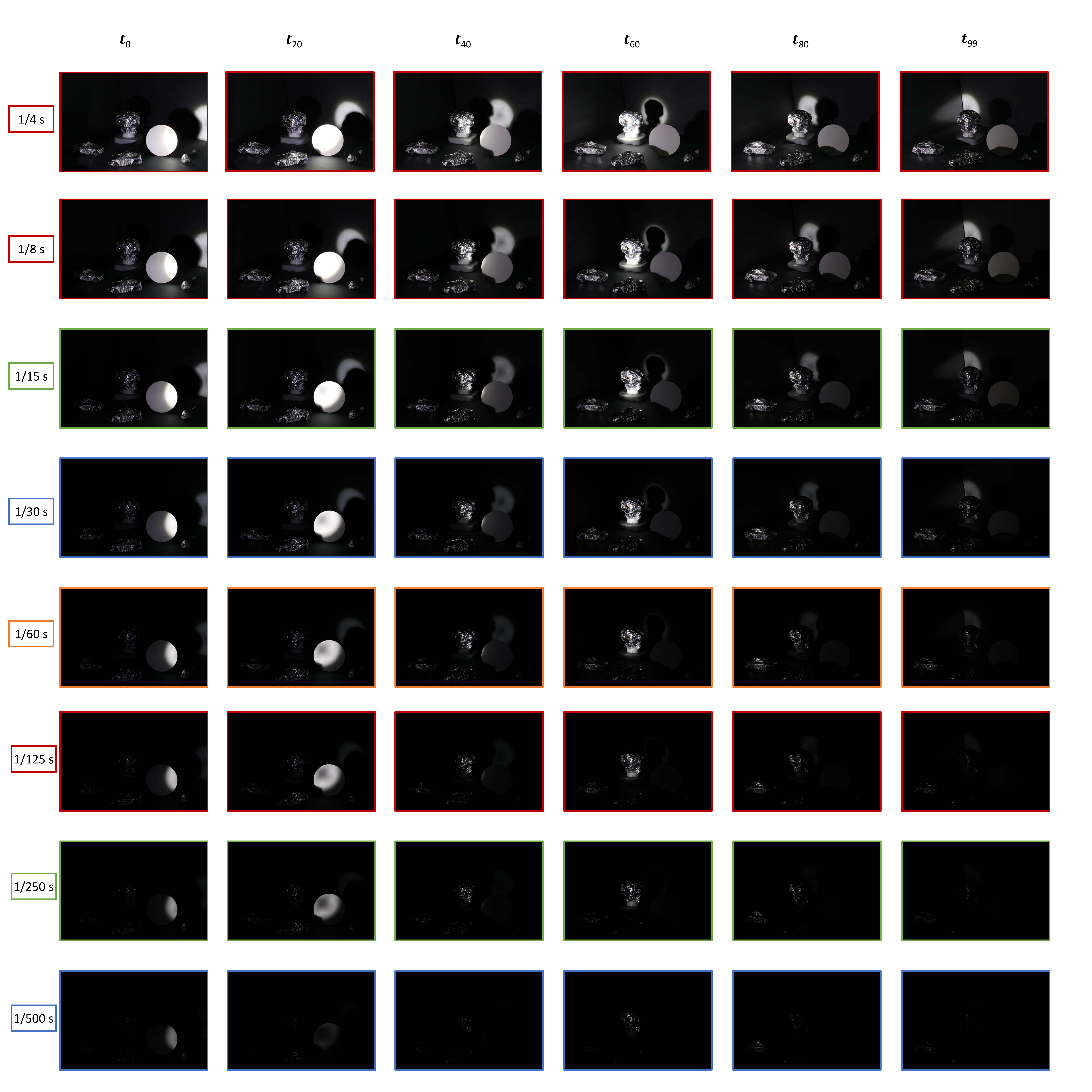}
	\end{center}
	\caption{Scene 7 exposure stack (\(\frac{1}{4}\)s - \(\frac{1}{500}\)s) at 6 different time steps.}
	\label{fig_data7_2}
\end{figure*}

\begin{figure*}
	\begin{center}
	\includegraphics[width=1\textwidth]{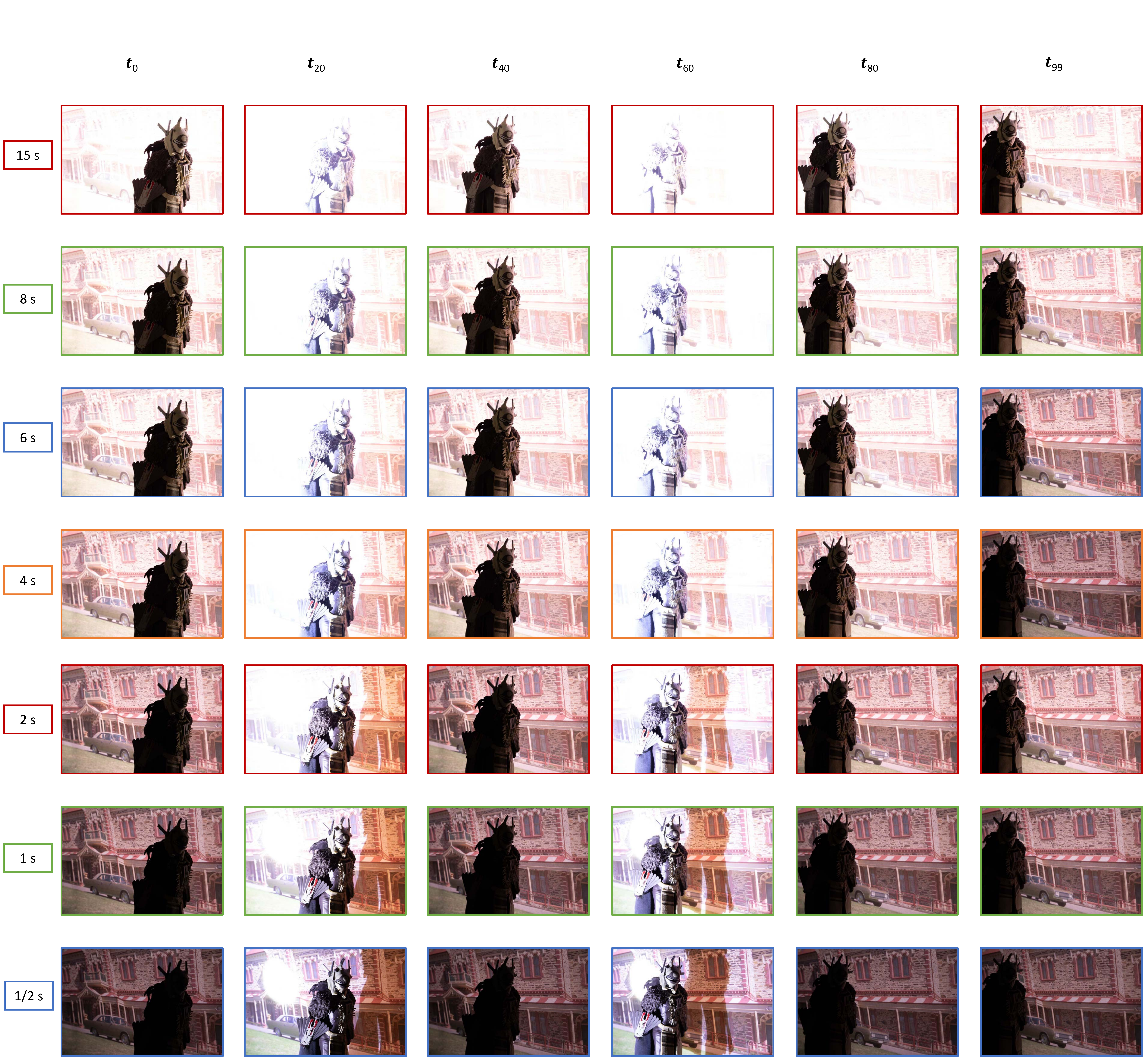}
	\end{center}
	\caption{Scene 9 exposure stack (15s - \(\frac{1}{2}\)s) at 6 different time steps.}
	\label{fig_data9_1}
\end{figure*}

\begin{figure*}
	\begin{center}
	\includegraphics[width=1\textwidth]{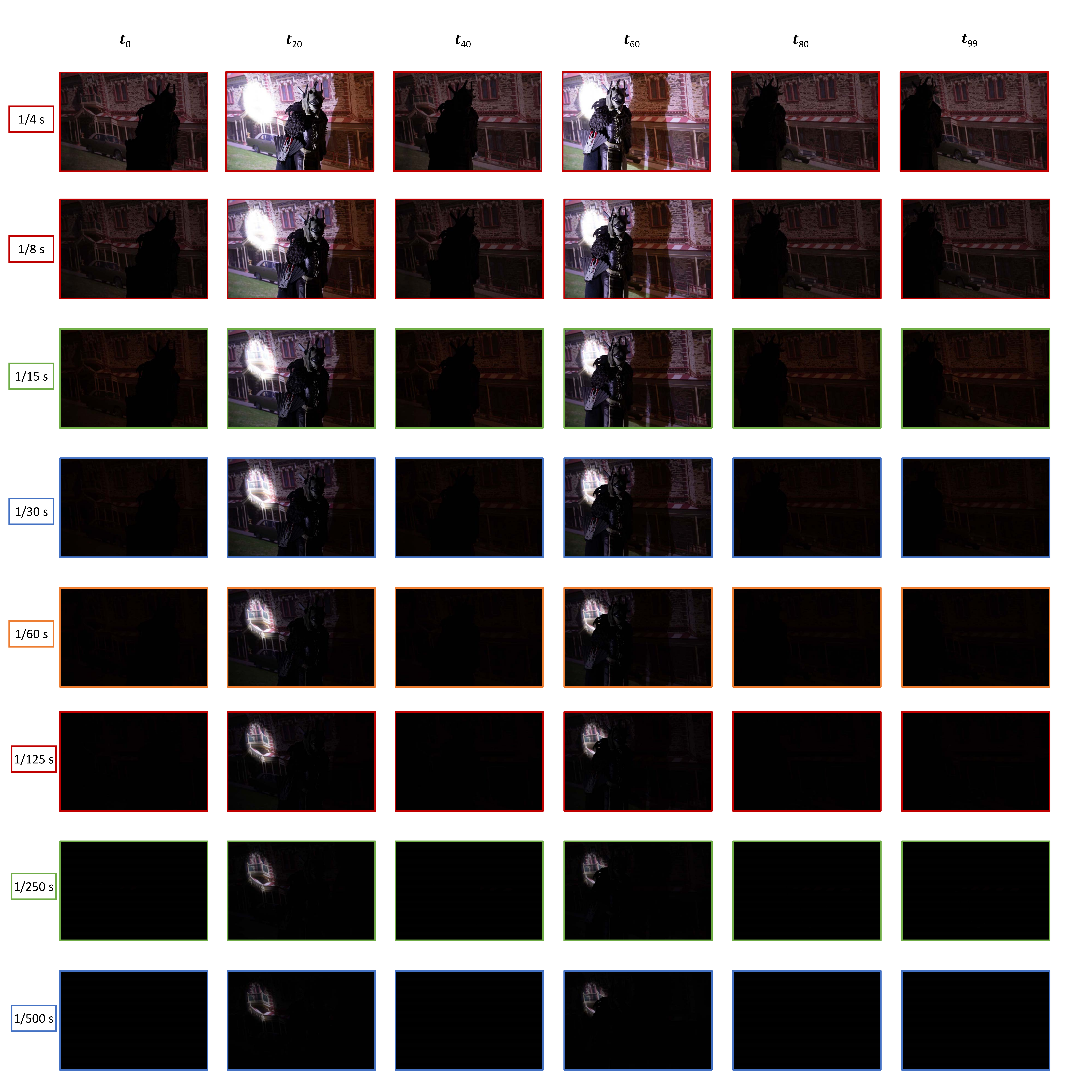}
	\end{center}
	\caption{Scene 9 exposure stack (\(\frac{1}{4}\)s - \(\frac{1}{500}\)s) at 6 different time steps.}
	\label{fig_data9_2}
\end{figure*}

Figures~\ref{fig_data1_1}-\ref{fig_data9_2} show examples of different scenes from our dataset; for each scene, we split the exposure stack across two figures. 





% \begin{figure*}
% 	\begin{center}
% 	\includegraphics[width=1\textwidth]{figures/Figures_dataset25_Part1.pdf}
% 	\end{center}
% 	\caption{}
% 	\label{fig_data8_1}
% \end{figure*}
% \begin{figure*}
% 	\begin{center}
% 	\includegraphics[width=1\textwidth]{figures/Figures_dataset25_Part2.pdf}
% 	\end{center}
% 	\caption{}
% 	\label{fig_data8_2}
% \end{figure*}

\end{multicols}
\end{document}